%% file: main.tex
\documentclass[twoside]{article}

\usepackage[accepted]{aistats2023}
\usepackage[utf8]{inputenc} 
\usepackage[T1]{fontenc}    
\usepackage{booktabs}       
\usepackage{amsfonts}       
\usepackage{amsmath,amssymb}
\usepackage{enumitem}
\usepackage[capitalize]{cleveref}
\usepackage{bbm}
\usepackage{caption}
\usepackage{subcaption}

\usepackage{standalone}

\usepackage{tikz}
\usetikzlibrary{fit}
\usetikzlibrary{arrows}
\usetikzlibrary{arrows.meta}
\usetikzlibrary{positioning}
\usetikzlibrary{matrix}

\usepackage{graphmod}
\usepackage{float}
\usepackage{url}

\usepackage[round]{natbib}

\input{m.macros}
\input{m.symbols}

\def\x{\vec{x}}
\def\z{\vec{z}}

\def\setX{\set{X}}
\def\setZ{\set{Z}}
\def\setU{\set{U}}

\def\fx{\psi^x}
\def\fxz{\psi^{xz}}
\def\fz{\psi^{z}}

\renewcommand\nn[1][n]{^{(#1)}}

\def\pthetaz{p_{\theta z}} 
\newcommand\pthetazxj[1][j]{f_{\theta #1}}
\newcommand\pthetazj[1][j]{F_{\theta #1}}

\def\pxemp{p_0}
\def\pxempj{p_{0j}}

\def\pxempjt{p_{0,jt}}
\newcommand\wtheta{W_\theta}

\newcommand\XN[1][N]{\ensuremath{\mathbb{X}\nn[#1]}}

\def\pthetaXN{\pr[\theta,\XN]}

\newcommand\etaz{\ensuremath{\eta_0}}
\newcommand\etaxjn[1][n]{\ensuremath{\eta_j(\x_j\nn[#1])}}
\newcommand\etaqn[1][n]{\ensuremath{\eta_q\nn[#1]}}

\newcommand\muz{\ensuremath{\mu_0}}
\newcommand\muxjn[1][n]{\ensuremath{\mu_j(\x_j\nn[#1])}}
\newcommand\muqn[1][n]{\ensuremath{\mu_q\nn[#1]}}
\newcommand\Vqn[1][n]{\ensuremath{V_q\nn[#1]}}

\newcommand\tletajn[1][n]{\ensuremath{\tilde\eta_j\nn[#1]}}

\def\eff{\mathcal{F}}

\begin{document}

%

%
\runningtitle{Unsupervised Representation Learning with Recognition-Parametrised Probabilistic Models}

\twocolumn[

\aistatstitle{Unsupervised Representation Learning with\\ Recognition-Parametrised Probabilistic Models}
\aistatsauthor{ William I. Walker*  \And Hugo Soulat* \And  Changmin Yu \And Maneesh Sahani* }
\aistatsaddress{Gatsby Computational Neuroscience Unit, University College London} ]

\begin{abstract}

We introduce a new approach to probabilistic unsupervised learning based on the \emph{recognition-parametrised model} (RPM): a normalised semi-parametric hypothesis class for joint distributions over observed and latent variables.
Under the key assumption that observations are conditionally independent given latents,
the RPM combines parametric prior and observation-conditioned latent distributions with non-parametric observation marginals.
This approach leads to a flexible learnt recognition model capturing latent dependence between observations, without the need for an explicit, parametric generative model. 
The RPM admits exact maximum-likelihood learning for discrete latents, even for powerful neural-network-based recognition.  We develop effective approximations applicable in the continuous-latent case. 
Experiments demonstrate the effectiveness of the RPM on high-dimensional data, learning image classification from weak indirect supervision; direct image-level latent Dirichlet allocation; and recognition-parametrised Gaussian process factor analysis (RP-GPFA) applied to multi-factorial spatiotemporal datasets.
The RPM provides a powerful framework to discover meaningful latent structure underlying observational data, a function critical to both animal and artificial intelligence. 

\end{abstract}

\section{INTRODUCTION}
Unsupervised representation learning plays a key role in systems that seek to learn and estimate
world state and latent structure from observations, including those that address real-world
reinforcement learning and robotics, tracking, semi-supervised task learning, and scientific
discovery.
Such systems have long been underpinned by the methods of probabilistic latent-variable modelling
\citep{bishop:2006,barber:2011,murphy:2022}.
In the most common approach, a model describes a family of distributions over a set of latent
variables and the conditional dependence of the observed variables on those latents.
Together, these define a directed acyclic graphical model (\cref{fig:rpmgraph}) or DAG.
Marginalising over the latents then results in a hypothesis class of joint distributions on the
observations.
Distribution parameters can be found by standard estimation techniques, identifying a model
within the class (or a posterior over models) that best matches the data distribution.


%
%
Although latent-variable models may also be used for sample simulation
\citep{goodfellow+al:2014:gan,kingma+al:2021:diffusion} or density estimation \citep{rezende+mohamed:2015:normflow}, they
play a key role in representation learning.
Many data sets exhibit complex dependence amongst observations, which arise
through common influences from unobserved but causally relevant features of the data generating
process.
By estimating models that render observations independent conditioned on latent state it is often
possible to tease out and represent such underlying features.
Indeed, it is this assumption of latent-conditioned independence---between the inputs to different
sensors, between sensor modalities, or between future and past---that provides the basis for
learning underlying structure in the absence of strong distributional assumptions.
%
\renewcommand*{\thefootnote}{}
\footnote{* These authors contributed equally}
\renewcommand*{\thefootnote}{\arabic{footnote}}
\setcounter{footnote}{0}

In this representation-learning view, the generative model serves to encode
structural priors about dependence and distribution, and the associated marginal
on observations underlies the choice of estimation objective, such as likelihood.
However, once learnt, neither is used directly.  Instead, the model structure and parameters are
used for inference or \emph{recognition}---to estimate the state of the world from sensory data
\citep{helmholtz:1867}.
This mismatch between the way the model is specified and how it is eventually used poses a challenge
to effective learning.
Generative models that are sufficiently complex, flexible and non-linear to parametrise real-world observations do not generally admit efficient tractable inference,
and so recognition is often approximated
\citep[e.g.][]{dayan+al:1995:helmholtz,jordan+al:1999:var,rezende+al:2014:vae,kingma+welling:2014:vae}.
These approximate methods lead to biases in parameter estimates \citep{turner+sahani:2011:ildn}, and
result in learnt representations that do not, in fact, match the learnt generative model.

Our goal here is to address such challenges to probabilistic representation learning.
We do so by introducing a form of semi-parametric model in which an explicit parametrisation of
the \emph{recognition} process is paired with a simplified non-parametric description of the
observations.
This \emph{recognition parametrised model} (RPM) defines a properly normalised joint distribution,
and thus (implicitly) a proper semi-parametric marginal distribution on observations.
Maximum-likelihood (ML) learning can be achieved exactly for models with discrete latent variables
(and a tractable internal graph), whilst in other settings it depends on potentially milder
approximations than do methods that pair recognition modelling with explicit generative models such
as the Helmholtz machine \citep{dayan+al:1995:helmholtz,vertes+sahani:2018:nips} or variational
autoencoder (VAE) \citep{rezende+al:2014:vae,kingma+welling:2014:vae}.
The RPM allows many different distributional and structural assumptions on the latent variables to
be combined with a recognition parametrisation, and pairs effectively with established techniques
such as variational message passing \citep{winn+al:2005:vmp} or variational Bayesian learning
\citep{attias:2000:vb} to estimate models with complex latent dependence.

Below, we first present a general formulation of the RPM (\cref{sec:rpm}), discuss inference and
learning in this general case (\cref{sec:ml}), and relate it to existing models
(\cref{sec:others}).  Thereafter we demonstrate the breadth of the framework by instantiating
different conditional structures and prior assumptions on the latent factors, and applying these to
appropriate data sets (\cref{sec:experiments}).

\section{THE RPM}
\label{sec:rpm}

\begin{table}%
  \hskip -1em
  \begin{tikzpicture}
    \matrix [matrix of nodes,row 3/.style={below,font=\tiny}] {
      Factor graph & DAG & RPM
      \\[-1ex]
      \vrule width0.3\columnwidth height1pt depth0pt 
      & \vrule width0.3\columnwidth height1pt depth0pt 
      & \vrule width0.3\columnwidth height1pt depth0pt 
      \\
      \setlength{\gmu}{4ex}
      \graphmod{graphical_models/generic}[factor.leftlabels]
      & 
      \setlength{\gmu}{4ex}
        \graphmod{graphical_models/generic}[dag][]
      &
      \setlength{\gmu}{4ex}
        \graphmod{graphical_models/generic}[rpm][]
      \\
      $\fz(\setZ)$  
      & $\pthetaz(\setZ)$   
      & $\pthetaz(\setZ)$
      \\
      $\fxz_j(\x_j, \setZ)$
      &$p_{\theta_j}(\x_j | \setZ)$
      & $\displaystyle\frac{\pthetazxj(\setZ | \x_j)}{\pthetazj(\setZ)}$
      \\
      $\fx_j(\x_j)$
      & 1
      & $\scriptstyle\pxempj(\x_j) = \frac1N\!\!\sum\limits_n \delta(\x_j - \x_j\nn)$
     \\
    };
  \end{tikzpicture}
  \vskip -2ex
  \caption{Conditional independence models and factor definitions.  Left column shows a generic factor graph with corresponding unnormalised factors.  Central column shows a directed graph. Right column shows the corresponding RPM factors.}
  \label{fig:rpmgraph}
\end{table}

Consider a set of observed (possibly vector-valued) random variables $\setX = \{\x_j : j = 1\dots
J\}$.  We seek to learn a model based on a set of underlying latent variables $\setZ = \{\z_l : l =
1\dots L\}$, given which the different $\x_j$ are conditionally independent.  These variables may be
loosely interpreted as causally relevant features responsible for the statistical interdependence of the
observations.  Our goal is to learn the joint distribution of the latents, along with a parametrised
model that infers a suitable belief over their values from observations.  We use the symbol $\pr{}$ (often with subscripts) to indicate complete (normalised) model distributions, and italicised symbols for factors within the models (\cref{fig:rpmgraph}), noting where these are individually normalised.  We write $\setX\nn = \{\x_j\nn : j = 1\dots J\}$ for the $n$th joint data observation, and $\XN = \{\setX\nn[1] \dots \setX\nn[N]\}$ for the entire set of $N$ observations.

The conditional independence assumption implies a factorisation (and corresponding factor graph)
\begin{equation} \label{eq:factor-one-Z}
   \pr{\setX, \setZ} \propto \fz(\setZ) \prod_j \fx_j(\x_j) \prod_j \fxz_j(\x_j, \setZ)  \,.
\end{equation}

In the RPM, these factors are parametrised as follows:
\begin{description}[labelindent=0em,leftmargin=0.5em,itemsep=0.2ex,topsep=0.2ex,partopsep=0pt]
  \item[$\fz(\setZ) \to \pthetaz(\setZ):$] a normalised distribution on the latent variables.  For multivariate \setZ\ this may itself be factored, with a corresponding latent graphical model.
  \item[$\fx_j(\x_j) \to \pxempj(\x_j):$]  a summary of the empirical marginal distribution of   each
    observed variable, with the property that it converges to the true distribution of $\x_j$ in the
    limit of infinite data.  In this paper we take $\pxempj(\x_j) = \frac 1N\sum_n \delta\lr[\big](){\x_j - \x_j\nn}$, the empirical measure with atoms at the $N$ data points $\x_j\nn$. However, the key definitions and results extend to alternatives, such as an adaptive kernel density estimate with kernel width that approaches 0 as $N$ grows, or (if known) a member of the true marginal distributional family specified
    by a sufficient statistic of the data.  The key is that $\pxempj$ is determined by the corresponding observations, with learning of the joint distribution focused on the other factors of the RPM.
  \item[$\fxz(\x_j,\setZ) \to \frac{\pthetazxj[j](\setZ| \x_j)}{\intdx[d\x_j] \pxempj(\x_j)
      \pthetazxj[j](\setZ | \x_j)}$] where $\pthetazxj[j](\setZ|\x_j)$ is a parametrised
    normalised distribution possibly, but not necessarily, defined on only a subset of the $\setZ$ (often on a
    single $\z_l$).  We write $\pthetazj(\setZ)$ for the mixture with respect to
    $\pxempj$  that appears in the denominator.  The numerator terms $\pthetazxj[j](\setZ| \x_j)$ will be referred to as \emph{recognition factors}.
\end{description}
Thus the full joint RPM model becomes
\begin{equation}\label{eq:joint}
   \pthetaXN{\setX, \setZ}= \pthetaz(\setZ) 
     \prod_j \lr[\Big](){
        \pxempj(\x_j)
        \frac{\pthetazxj(\setZ| \x_j)}{\pthetazj(\setZ)}
    }  \,,
    \end{equation}
where the observed dataset $\XN$ appears in the subscript to
emphasise that the model parametrisation itself depends on the data through $\pxempj$ and so $\pthetazj$.

With this choice of parametrisation we have
\begin{align}
\pthetaXN{\setZ}
  &=  \prod_j \intdx[d\x_j] \left(
     \frac{\pxempj(\x_j) \pthetazxj[j](\setZ| \x_j)}{\pthetazj(\setZ)}
     \right) \pthetaz(\setZ)  \nonumber \\
  &=\pthetaz(\setZ)\,,
\end{align}
so that the parametrised factor on the latents corresponds to the prior distribution implied by the
joint (as is also the case for a DAG).  This result confirms that the RPM is properly normalised.
%
%
The posterior
\begin{multline}\label{eq:recog-explicit-integral}
  \pthetaXN[\big]{\setZ | \setX\nn}
  \propto \prod_j 
        \frac{\pthetazxj[j](\setZ| \x_j\nn)}{\intdx[d\x_j] \pxempj(\x_j) \pthetazxj[j](\setZ | \x_j)}
    \pthetaz(\setZ) \\
  = \frac 1{\wtheta(\setX\nn)} \prod_j \pthetazxj[j](\setZ| \x_j\nn) \;
    \frac{\pthetaz(\setZ)}{\prod_j  \pthetazj[j](\setZ)}
\end{multline}
can be found by normalising the product of learnt factors.
The normaliser $\wtheta(\setX)$ also gives the relative
density of the implicit RPM joint on the observed variables, $ \pthetaXN{\setX} = \prod_j \pxempj(\x_j) \wtheta(\setX)$.  The joint is supported on the
Cartesian product of the supports of $\pxempj(\x_j)$---an irregular grid of atoms for atomic $\pxempj$ as we assume here.

\paragraph{Exponential-Family Parametrisation}

Beyond the need for normalisation, the factors $\pthetaz$ and $\pthetazxj(\cdot|\x_j \nn)$ in the RPM as defined above can be chosen freely.  In practice, we will often assume that they lie within a common exponential family with combined sufficient statistic $t(\setZ)$ defined on all the latent variables, and log-normaliser $\Phi$. The prior factor will be taken to have natural parameter $\etaz$.  The natural parameters of the recognition factors are now parametrised functions of the observations given by $\etaxjn$ (which may be constrained to have constant outputs along some dimensions when the recognition factors target a subset of the latent variables). See also \cref{tab:tnotations}.

With these choices (and assuming uniform base measure for $\pthetaz$), we can write the implied \emph{generative} conditionals of the RPM as:
\def\TZ{t(\setZ)}
\def\xj{\x_j}
\def\etaxj{\eta_j(\x_j)}
\def\chixj{\chi_j}
\def\Phixj{\Phi_{\x_j}}
\begin{align*}
  \pr[\theta,\XN]{\xj|\setZ} 
   &
     = \pxempj(\xj) \frac{e^{\etaxj\tr\TZ - \Phi(\etaxj)}}{\pthetazj(\setZ)}
\\  
    &= \chixj(\xj)\, e^{\TZ\tr \etaxj - \Phixj(\TZ)}
\end{align*}
with
\begin{align*}
\chixj(\xj)
    &= \frac1C \; \pxemp(\xj) e^{-\Phi(\etaxj)}  
\intertext{for constant $C$ and}
  \Phixj(\TZ) &= \log \intdx[d\xj] \frac1C\; \pxemp(\xj) e^{- \Phi(\etaxj)} e^{\etaxj\tr\TZ}
  \\
  &= \log \intdx[d\xj] \chixj(\xj) e^{\etaxj\tr\TZ}\,.
\end{align*}
Thus, this form of RPM induces an exponential family conditional on each $\x_j$ in which the parameters of $\etaxj$ determine both the sufficient statistic and the base measure, the latter also depending on the observed marginal.  This expression underlines the expressiveness of the RPM with flexibly parametrised recognition factors.

%

\section{MAXIMUM-LIKELIHOOD LEARNING}
\label{sec:ml}

\subsection{Variational Free Energy}

As is the case for other latent-variable models, ML estimation in the RPM can be achieved using the
Expectation-Maximisation (EM) algorithm and related methods.  We adopt the viewpoint of
\citet{neal+hinton:1998} and frame EM as coordinate ascent of a variational free energy 
derived by applying Jensen's inequality to the log likelihood:
\begin{multline}
    \sum_n \log \pthetaXN{\setX\nn} \ge
    \eff(\theta, q(\{\setZ\nn\}))   \nonumber
    \\
     = \angles[\Big]{\sum_n\log\pthetaXN{\setX\nn,\setZ\nn}}
      + \entropy{q}
\end{multline} 
where angle brackets indicate expectations over the variational distribution $q$ and
$\entropy{\cdot}$ is the entropy. Dropping the term in $\pxempj$ which is independent of $\theta$ and $q$, $\eff$
can be written in terms of Kullback-Leibler (KL) divergences as
\begin{multline} \label{eq:eff-klform}
    \hskip -1em - \eff(\theta, \{q\nn(\setZ\nn)\}) 
    \underset{+C}{=} \sum_n  
        \KL[\big]{q\nn}{\pthetaz} \\
       + \sum_{nj}
          \KL[\big]{q\nn}{\pthetazxj(\cdot | \x_j\nn)}
       - \sum_{nj}
         \KL[\big]{q\nn}{\pthetazj}
    \,,
\end{multline}
where we have used the fact that the optimal $q$ has the form $\prod_{n}
q\nn(\setZ\nn)$, and the distributions in the latter KL divergences range over only the $\z_l$ that are targeted by the corresponding $\pthetazxj$ or $\pthetazj$.

Alternating maximisation of $\eff$ with respect to $q$ (the ``E-step'') and $\theta$ (``M-step'')
will converge to a (possibly local) mode of the likelihood, provided that each maximum can be
achieved.  This is straightforward in cases where the latent targets of each $\pthetazxj$ are
discrete-valued variables (so that $\pthetazj(\z_k) = \intdx[d\x_j] \pxempj(\x_j)
\pthetazxj(\z_k|\x_j)$ is an easily computed discrete distribution) and $\pthetaz$ has conjugate
structure and sufficiently small junction tree width to be computationally tractable.  Examples of
such exact ML learning in an RPM are explored below in \cref{sec:peer,sec:lda}.

\subsection{E-step for Continuous-Valued Latent Variables}
\label{sec:ml:estep}

The situation is more complex when the latent-variable targets of $\pthetazxj$ are continuous-valued.
Even assuming that the graphical structure and factor potentials that compose $\pthetaz(\setZ)$
allow tractable marginalisation, and that the terms $\pthetazxj[j](\setZ | \x_j)$ provide conjugate
factors, the inverse expectation factors $\lr(){\intdx[d\x_j] \pxempj(\x_j) \pthetazxj[j](\setZ |
  \x_j})\inv$ will generally break conjugacy and thus analytic tractability.
%
%
A natural approach in this case is to constrain $q\nn$ to live within the conjugate family defined
by $\pthetaz$ and $\pthetazxj[j](\cdot | \x_j)$.
This constraint renders the first two KL divergences in \cref{eq:eff-klform} tractable, but requires
approximation to evaluate the third.
However, by contrast to the analogous standard parametric variational assumption made in the context
of non-conjugate parametrised \emph{generative} models (e.g. the VAE) the impact of the constraint
in the RPM may be negligible in the large-data in-model limit.
Specifically, if the true posterior on $\setZ$ lies within the parametric class of $\pthetazxj$ then,
in the limit of large data one potential set of ML parameters will be such that $\pthetazj(\setZ) =
\intdx[d\x_j] \pxempj(\x_j) \pthetazxj(\setZ|\x_j) \to \pthetaz(\setZ)$.
This implies that the penalty of \emph{assuming} that $\pthetazj(\setZ)$ has the exponential family form will
become negligible in the large data limit for this in-model conjugate case.

There are at least three approaches to optimising $q$ in the continuous case.  We consider the exponential family parametrisation, and constrain
$q(\setZ\nn)$ to be in the same family, with natural parameter $\etaqn$.  We will sometimes also
require the moment parameters of the various distributions (i.e.\ the expectations of $t(\setZ)$ under the corresponding natural parameter).
These will be written $\muz$, $\muxjn$ and $\muqn$ for the prior, recognition factors and $q\nn$ respectively.  See \cref{tab:tnotations}.

\paragraph{Reparametrised Monte-Carlo} The first approach adopts a strategy used extensively in the
VAE literature where it is known as ``reparametrisation''. It is often possible to express a sample
from the exponential family of interest as a parametrised function of a sample from a fixed
distribution.  A common example is the normal family, where samples from $\normal{\vec\mu}{\Sigma}$
can be expressed in terms of $\vec\epsilon_i \sim \normal{0}{I}$ as $\Sigma^{\half}\vec\epsilon_i +
\vec\mu$.  This allows a Monte-Carlo estimate of the expectation of $\pthetazj$ to be written as a
function of the parameters $\etaqn$ and a fixed set of samples $\{\vec\epsilon_i\}$.  The other
expectations can be evaluated analytically under our conjugacy assumptions. Thus, it becomes
possible to optimise $\eff$ with respect to $\etaqn$ by gradient ascent.   
%
While accurate with large numbers of samples, this approach may be computationally expensive for high-dimensional problems.

\paragraph{Second-order Approximation} An efficient approximation of $\angles{\log \pthetazj} $ can
be obtained by generalising an approach taken by \citet{braun2010variational}.  We expand $\log
\pthetazj(\setZ)$ to second order in $t(\setZ)$ around its expectation  $\muqn$ under the variational distribution $q\nn$.
Then, writing $\Vqn$ for the variance of $t(\setZ)$ under $q\nn$, we have
\begin{multline}
  \hskip -1em
  \angles[\big]{ \log \pthetazj(\setZ)} \approx
  \log \frac{1}{N} \sum_{m=1}^N e^{\etaxjn[m]^\top \muqn - \Phi(\etaxjn[m])}
   \\
   + \frac12 \text{tr} \left( \vec{\eta}_j^\top
   \Vqn \vec{\eta}_j \left[ \text{diag}(\vec{\pi}_j\nn) - \vec{\pi}_j\nn{\vec{\pi}_j\nn}^\top
     \right] \right)
\end{multline}
where  $\vec{\eta}_j =\lr[\big][]{\eta_j(\x_j\nn[1]), \dots, \eta_j(\x_j\nn[N])}$ and 
$\vec{\pi}_j\nn$ is an $N$-dimensional vector with components  (for $m=1\dots N$)
\begin{equation}
    \pi_{mj}\nn = \frac{e^{ \etaxjn[m]\tr \muqn  - \Phi(\etaxjn[m])}}{\sum_{p} e^{
        \eta(\etaxjn[p])\tr \muqn - \Phi(\etaxjn[p])}}\,.
\end{equation}

This approximation no longer guarantees a lower bound on the free energy but we demonstrate its efficacy in practice.

\paragraph{Interior Variational Bound} A third approach introduces
auxiliary variational parameters to obtain a second bound on $\eff$.  Focusing on the
$\pthetazj$-dependent terms as above (and again using angle brackets for expectations
under $q$) we introduce functions $\tilde f_j\nn(\setZ)$ and use Jensen's inequality to write
\begin{multline}\label{eq:dvbound}
 \hskip -1em 
 \angles{\log \frac{\pthetazxj(\cdot| \x_j\nn)}{\pthetazj}}
    \ge \angles{\log \frac{\pthetazxj(\cdot| \x_j\nn)}{\tilde f_j\nn q\nn}}
          - \log \angles{\frac{\pthetazj}{\tilde f_j\nn q\nn}}
  \\
    = - \KL[\big]{q\nn}{\pthetazxj(\cdot| \x_j\nn)}
          - \angles[\big]{\log \tilde f_j\nn} 
          - \log \Gamma_j\nn \,,
\end{multline}
where $\tilde \Gamma_j\nn = \intdx[d\setZ] \pthetazj(\setZ) \big/ \tilde f_j\nn(\setZ)$.  Inserting this
bound into \cref{eq:eff-klform} and rearranging gives
\begin{multline}\label{eq:efftilde-klform}
\hskip -1em
  \widetilde \eff
    = \sum_n \log\pr[\theta,\XN][\big]{\setX\nn} 
    - \sum_n \KL[\big]{q\nn}{\pr[\theta,\XN][\big]{\!\cdot\!|\setX\nn}}
\\    
    - \sum_{nj} \KL[\Big]{q\nn}{\frac1{\tilde\Gamma_j\nn} \frac{\pthetazj}{\tilde f_j\nn}}
\end{multline}
with $\widetilde\eff\lr(){\theta, q, \{\tilde f_j\nn\}} \le \eff(\theta,q)$ lower-bounding the conventional free energy.  
%
If we now choose $\tilde f_j\nn(\setZ) = \exp\lr[\big](){t(\setZ)\tr \tletajn}$ with the constraint that $\etaxjn[m] - \tletajn$ is a valid natural parameter for all $(m, n)$, then the right hand-side of \cref{eq:dvbound} is closed-form.
Furthermore, if $\pthetazj$ approaches the exponential family with statistic $t(\setZ)$, as in the in-model conjugate case, then it will be possible to choose $\tletajn$ to set the final KL-divergence in \cref{eq:efftilde-klform} close to 0, restoring a tight bound.

\subsection{M-step}
\label{sec:ml:mstep}

The generalised M-step of EM increases the free energy with respect to the parameters $\theta$ while holding the variational distribution $q$ fixed \citep{neal+hinton:1998}.  For many RPMs, the parameter vector will divide into disjoint subsets that determine $\pthetaz$ and the $\pthetazxj$ (possibly shared for multiple $j$).  In this case, the update for the $\pthetaz$ group will be broadly identical to the usual EM update.  For $\pthetazxj$ the update requires gradients of both $\angles[\big]{\log\pthetazxj(\setZ\nn|\x_j\nn)}$ and $\angles{\log\pthetazj(\setZ\nn)}$.  

For discrete-valued latent variables where the E-step is exact, the corresponding M-step is straightforward, possibly incorporating backpropagation of gradients where $\pthetazxj$ has neural network form and amenable to automated gradient-based optimisation.
For continuous-valued latent variables the corresponding step depends on the E-step approach used.  Reparametrisation and the second-order approximation both provide an explicit estimate of $\eff$ which can be increased directly.  When employing the interior variational bound, we instead increase the term $\widetilde\eff$ (see Appendix).  

It is worth noting that, although the discussion above has focused on cases where the latent distribution $\pthetaz$ is tractable (in the sense that the marginals needed for learning can be computed efficiently) the RPM can also be seamlessly combined with standard approximate variational inference and learning methods.  This includes variational Bayesian methods to obtain approximate posteriors on parameters.

\section{RELATIONSHIPS TO OTHER MODELS}
\label{sec:others}

\paragraph{Dual Generation-Recognition Models}  Many archetypal learning architectures for latent-variable models,
including the variational autoencoder \citep[VAE;][]{kingma+welling:2014:vae} and Helmholtz machine
\citep{dayan+al:1995:helmholtz}, employ parametrised recognition networks in support of learning an explicit generative
model for data.
The associated objective functions are usually derived from the likelihood of the generative parameters, with the
recognition model supplying `E-step' inference in an EM-like approach \citep{neal+hinton:1998,jordan+al:1999:var}.
However, the true posterior distribution
over the latents implied by the generative structure rarely lies within the class of functions described by the
recognition model parametrisation.
This mismatch induces an intrinsic bias in the estimates of the generative parameters
\citep[e.g.][]{turner+sahani:2011:ildn}, which can be seen either as a necessary compromise or (for the VAE) as a
reframing of the objective function from the likelihood to the variational lower bound \citep{jordan+al:1999:var}.

Recent work has sought to lessen the bias by introducing a more flexible posterior representation
\citep{rezende+mohamed:2015:normflow,vertes+sahani:2018:nips,wenliang+al:2020:icml}, or tighter variational bounds than the classic free energy form
\citep{burda+al:2016:iwae,maddison+al:2017:fivo,masrani+al:2019:tvo}.
However, these extensions retain the emphasis on approximate ML estimation of a parametric \emph{generative} process
with a specific noise model, potentially guiding the latent representation towards details of individual data elements
that may not be representationally useful.
By contrast, the RPM likelihood emphasises latent structure that captures dependence between data elements, dispensing
with a parametrisation of the marginal distributions of individual elements and corresponding noise.  Intuition suggests
that this joint structure is most likely to reflect latent ``causal'' elements, and so may be most valuable for
decision making.
Furthermore, although approximation is necessary for RPM models with continuous-valued latent variables, the impact of the approximation will not always persist as the data set grows (see the discussion of in-model conjugacy in \cref{sec:ml:estep}).

The RPM is also directly compatible with graphical (i.e. conditional-independence-based) prior structure within the
latents, as explored in various models below.  Analogous structured versions have been explored in the context of
generation-recognition parametrisations; but complications arise from the need to backpropagate gradients through
message passing in the latent graph of structured VAEs \citep{johnson+al:2016:svae} or from the need to approximate message passing in
complex Helmholtz machines \citep{vertes+sahani:2019:nips,wenliang+sahani:2019:nips}.

\paragraph{Undirected Models}  Latent-variable models may also be parametrised in a factored form corresponding to an
undirected graph, exemplified by the Boltzmann machine \citep{ackley+al:1985:boltzmann}.
Factor models with observations conditionally independent given the latents \emph{and vice versa} (such as the
restricted Boltzmann machine \citep[RBM;][]{smolensky:1986:rbm,hinton:2002:cd} or exponential-family harmonium
\citep{welling+al:2004:expfamharm}) may be viewed as restricted and unnormalised variants of the RPM.
Inference follows directly from the parametric form, but only because the latents are also conditionally independent
given the observations.
In other words, whereas the RPM can incorporate factors that link \emph{all} the latents to each observation separately
(see \cref{eq:factor-one-Z}), the RBM is restricted to pairwise factors linking individual latents and observations, or
more generally factors that link disjoint subsets of each.
Furthermore, the marginal prior on the latents is implicit and typically inaccessible and ML learning requires sampling
from the model, most often by Markov-chain methods.
Again this contrasts with the efficient ML learning of the RPM.

\paragraph{Noise-contrastive Estimation and InfoNCE}  An alternative to ML estimation is often applied to
``energy-based'' models, where an unnormalised data density is expressed as a parametrised non-negative function of the
observations (the logarithm of this function is the ``energy'').  The idea behind noise-contrastive estimation \citep{gutmann+hyvarinen:2010:nce} is
to train the energy as though it were the log-odds of a classifier that seeks to distinguish genuine observations from
corrupted ones.  This makes sense because, in the large-data limit, the optimal log-odds values correspond to the ratio
of the model log-likelihood on the genuine data to that on the corruptions.  One common form of corruption is to break
each observation into two components and shuffle these components around.  In this case, known as InfoNCE \citep{oord+al:2018:cpc}, the log-odds-like cost function approaches the mutual information between the components. 

Recall that the RPM data measure is defined on the cross-product of the empirical marginal summaries $\pxempj(\x_j)$,
weighted by $W_\theta(\setX)$.  RPM learning can thus be viewed as a process of maximising weights on the observations, which---as the distribution is normalised---must come at the expense of weights elsewhere.  Thus,
with the empirical delta-function measure, the RPM also learns to contrast real observations from shuffled versions.
Indeed, this link between InfoNCE and a probabilistic model has been noted previously \citep{aitchison:2023:ssvae}, though
the model proposed there was based on the (unknown) true data marginals rather than the empirical measures, and so
remained intractable.

The ``shuffling'' in the RPM is implicit, and involves all $J$ conditionally independent observed variables rather than
just pairs.  Furthermore, the normalised latent variable formulation (missing in energy-based approaches) provides
access to efficient message-passing inference in complex models, as well as to variational and other well-developed
tools of learning in probabilistic graphical models.  And the learnt recognition model provides proper posterior beliefs
over latent variables which can, as argued above, form the basis of optimal Bayesian decision making.

\section{EXPERIMENTS}\label{sec:experiments}
We demonstrate the flexibility of the RPM on a range of discrete- and continuous-latent problems:  weakly supervised categorisation, a pixel-level extension of Latent-Dirichlet Allocation (LDA) \citep{blei2003latent} to images, and non-linear recognition-parametrised Gaussian Process Factor Analysis (RP-GPFA) \citep{yu2008gaussian, duncker2018temporal}. 

The RPM performance was compared to that of appropriate VAEs to provide the most appropriate baseline.  Both VAE and RPM are normalised probabilistic models where we could equate distributional assumptions and recognition architecture.  
In all experiments the training data, prior and recognition architecture were identical for RPM and VAE.  The only differences were in the generative model that had to be instantiated for the VAE (which was set to an artificial neural network plus noise), and the corresponding learning algorithms.  Derivations and details are provided in the appendices. A comparison of compute time for two of the experiments is shown in \cref{fig:timing}.

\subsection{Peer Supervision}
\label{sec:peer}
\begin{figure}
\begin{minipage}{0.2\linewidth}
\begin{center}
\vspace{0.65cm}
  {\scriptsize\setlength{\gmu}{5ex}
   \graphmod{graphical_models/peer_supervision_gen}[genpeer][]}\\  
\vspace{0.15cm}
\footnotesize  Generative \\ Model
\end{center}
\end{minipage}
\begin{minipage}{0.5\linewidth}
\begin{center}
\vspace{0.5cm}
\includegraphics[width= 1 \linewidth, trim= 0cm 0cm 0cm 0cm, clip]{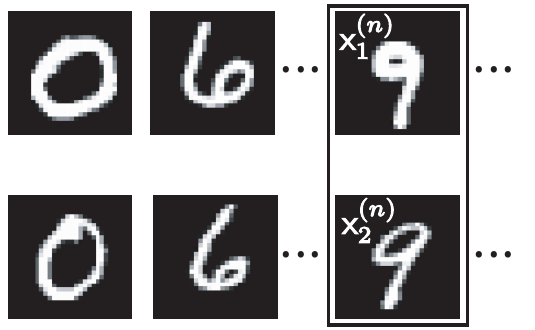} 
\footnotesize  Observations \hspace{0.3cm} $\setX \nn$
\\ \hspace{0.0cm}
\vspace{0.0cm} 
\end{center}
\end{minipage}
\begin{minipage}{0.2\linewidth}
\begin{center}
\vspace{0.2cm}
  {\scriptsize\setlength{\gmu}{5ex}
   \graphmod{graphical_models/peer_supervision_rpm2}[rpmpeer][]}\\ 
\footnotesize   RPM \\ Parametrisation
\end{center}
\end{minipage}\label{fig:FAgenerative}\hfil
\vspace{0.2cm}
        \caption{Peer-supervised learning. Each pair of observations $\setX = \{ \x_1, \x_2\}$ is conditionally independent given their shared digit identity $\z$.}
    \label{fig:Peer-Supervision}
\end{figure}

In the first experiment, the observations $\x_j$ are groups of MNIST \citep{deng2012mnist} images representing $J$ (here $2$) different renderings of the same digit. The data set is structured in this way so that the $J$ images are conditionally independent given the (unknown) digit identity.  Thus, we expect the RPM to extract identity without explicit label information -- a setting we term ``peer supervision'' (\cref{fig:Peer-Supervision}). The RPM is constructed with a single discrete-valued latent $\z$, and a recognition network (two convolutional layers, pooling, two linear layers and rectified linear activation function (ReLU) trained using Adam) with shared parameters $\theta$ for both factors.  The learned recognition network achieved an average test set classification accuracy of $0.87 \pm 0.09$ over different random seeds, achieving an accuracy of $0.96$ on 4 out of 10 runs.
Failures occurred predominantly when multiple (usually two) digits were systematically mapped to the same latent---a phenomenon possible in the absence of explicit label supervision.
This effect results in a correlation between the average posterior entropy and the classification accuracy (see \cref{fig:peer-sup-confusion}). 
%
When comparing the recognition network accuracy on MNIST test set, RPM outperforms both Vector Quantised-VAE  and VQ-VAE trained using the Gumbel Softmax categorical reparametrisation (GS-Soft VQ-VAE) \citep{sonderbypoole2017, van2017neural, maddison2016concrete}  See \cref{tab:baseline}. Implementation details can be found in \cref{sec:peer_super_supplement}.
%

\begin{table}
\small
\begin{center}
\begin{tabular}{cccc}
\textbf{VQ-VAE} & \textbf{GS-VAE} & \textbf{GS-S VQ-VAE} & \textbf{RPM} \\
\hline
$0.26 \pm 0.25$  &$0.46 \pm 0.06$ & $0.77 \pm 0.04$ & $\mathbf{0.87 \pm 0.09}$\\ 
\end{tabular}
\end{center}
\caption{Accuracy (higher better) on test MNIST data of recognition networks trained by peer supervision.}
\label{tab:baseline}
\end{table}


\subsection{RP-LDA}
\label{sec:lda}
\begin{figure}
  \begin{minipage}{0.49\linewidth}
    \footnotesize 
    \begin{flushleft}
      \hspace{0.3cm}
      (a) Generative Model
     \end{flushleft}
    \begin{center}
      \resizebox{!}{4.5cm}{
        \footnotesize\setlength{\gmu}{5ex}
        \graphmod{graphical_models/lda}[][]
      }
    \end{center}
  \end{minipage}
  \begin{minipage}{0.49\linewidth}
  \footnotesize 
  \begin{flushleft}
    \hspace{0.0cm}
    (b) RPM Parametrisation
  \end{flushleft}
  \begin{center}
    \resizebox{!}{4.5cm}{
      \setlength{\gmu}{5ex}
      \graphmod{graphical_models/lda}[rpm][]
    }
  \end{center}
\end{minipage}
\includegraphics[width= 1 \linewidth, trim= 0cm 0cm 0cm 0cm, clip]{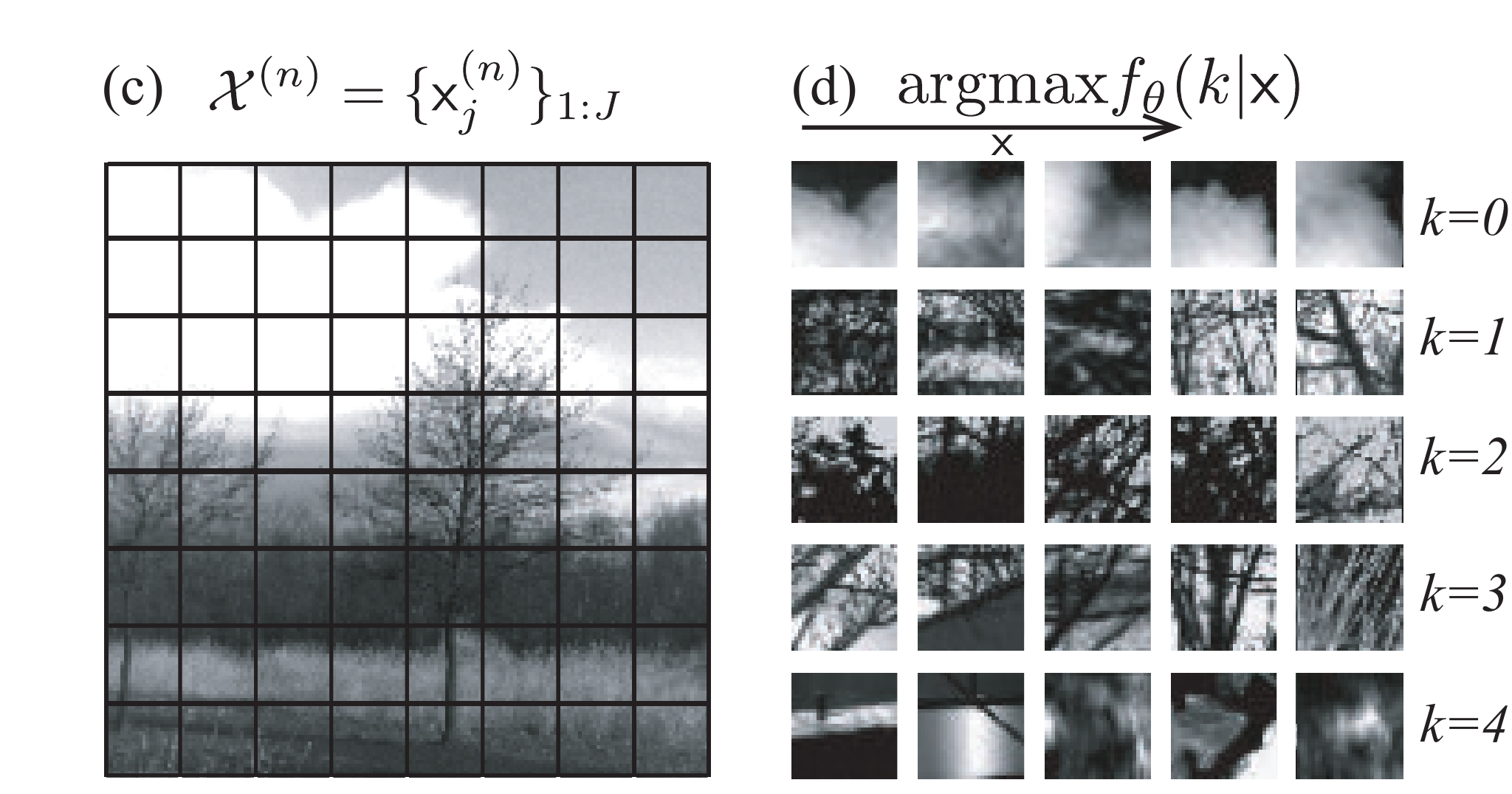}
\centering
\caption{\label{fig:RP-LDA} (a) Latent Dirichlet Allocation and (b) RP-LDA. (c) Splitting of an image in patches. (d)  Five most representative patches for five texture categories sorted using the recognition network outputs.}
\end{figure}

A second RPM instance builds on latent Dirichlet allocation (LDA) models and variational Bayes to identify pixel-level statistical regularities corresponding to textural properties of image subpatches (\cref{fig:RP-LDA}). 
Images, indexed by $n$, are decomposed into $J$ smaller non-overlapping sub-patches $\x_{j}\nn$ (\cref{fig:RP-LDA}c) which are assumed to be conditionally independent given a discrete latent texture identity $\z_j\nn$. 
The $\z_j\nn$ are drawn from random categorical distributions $\omega\nn$, which each gives the distribution of textures in the corresponding image, and is in turn drawn from a Dirichlet prior (with uniform parameter $\alpha$).
A recognition network (with the same structure as in \cref{sec:peer}) is shared across patches, and outputs a categorical distribution over texture identities given the patch pixel values. Writing $\setZ =  \{ \z_j: j=1\dots J\} \cup \{ \omega \}$, RP-LDA takes the form
\begin{equation}
\hfill
   \pr{\setX, \setZ}=  \prod_{j=1}^J 
     \pxempj(\x_j)
     \frac{ f_\theta(\z_j| \x_j)}{F_\theta(\z_j)}
      p \left(\z_j | \omega \right) p \left(\omega | \alpha \right).
\end{equation}

Applied to images from the van Hateren database \citep{van1998independent}, RP-LDA recovers textural components (clouds, branches, pavements, etc.).  \cref{fig:RP-LDA}d shows representative patches $\x_j$ that are most robustly assigned to a single textural category. Further details and derivations are given in \cref{sec:supp-lda}. 
%


\subsection{RP-GPFA}

Finally, we model continuous multi-factorial temporal dependencies by introducing recognition-parametrised Gaussian process factor analysis (RP-GPFA; \cref{fig:RP-GPFA}). We consider $J$ observed time-series measured over $T$ timesteps: $\setX = \{\x_{jt} : j = 1\dots J,\; t=1\dots T\}$.
We seek to capture both spatial and temporal structure in a set of $K$-dimensional underlying latent time-series $\setZ = \{\z_t : t = 1\dots T\}$, such that the observations are  conditionally independent across series and across time.  The RPM thus takes the form
\begin{equation}
   \pthetaXN{\setX, \setZ}= 
   \prod_{j=1}^J
   \prod_{t=1}^T \lr(){
     \pxempjt(\x_{jt})
     \frac{\pthetazxj(\z_t| \x_{jt})}{\pthetazj(\z_t)}
     } \pthetaz(\setZ) \,.
\end{equation}

%

The prior on $\setZ$ comprises independent Gaussian Process priors over each latent dimension $\setZ_k = \{z_{kt} : t = 1\dots T\}$
\begin{equation}
\label{eq:GPprior}
    \pthetaz(\setZ) = \prod_{k=1}^{K} p_k(\setZ_k) \,;\quad
    p_k(\cdot) = \mathcal{GP}(0, \kappa_k(\cdot,\cdot)) \,.
\end{equation}
The recognition factors are parametrised by a neural network with weight $\theta_j$ that outputs the parameters of a multivariate normal distribution, i.e. $\pthetazxj(\z_t | \x_{jt}) = \normal{\z_t ; \mu_\theta [\x_{jt}]}{\Sigma_\theta [\x_{jt}]}$.
%
%
We use the sparse variational GP approximation \citep{titsias2009variational} to improve scalability. The model is augmented with $M$ inducing points (IP) for each latent dimension ($k=1\dots K$) and each observation ($n=1\dots N$). This smaller set ($M < T$) of fictitious measurements is optimised to efficiently represent function evaluations. For simplicity, IP are defined at fixed and shared locations.  We restrict our experiments to using the radial basis function kernel, but the method can accommodate any GP prior. 

\begin{figure}
\begin{minipage}{0.49\linewidth}
\begin{center}
\vspace{0.45cm}
  {\tiny\setlength{\gmu}{5ex}
   \graphmod{graphical_models/gpfa_gen}[gen][]}\\  
\vspace{0.45cm}
\footnotesize   Generative Model 
\end{center}
\end{minipage}
\begin{minipage}{0.49\linewidth}
\begin{center}
  {\tiny\setlength{\gmu}{5ex}
   \graphmod{graphical_models/gpfa_rpm}[rpm][]}\\  
\footnotesize   RPM Parametrisation
\end{center}
 \end{minipage}
     \caption{RP-GPFA}
    \label{fig:RP-GPFA}
\end{figure}

\subsubsection*{Performance range}
\Cref{tab:BallBaseline} gives the median and inter-quartile range for each of the RP-GPFA experiments described in the main text.
\begin{table*}
\begin{center}
\small
\begin{tabular}{ccccccc}
\multicolumn{2}{c}{\textbf{ }} & \multicolumn{2}{c}{\textbf{sGP-VAE}} & \multicolumn{3}{c}{\textbf{RP-GPFA}}\\
& & 1D & 2D & Monte-Carlo & 2\textsuperscript{nd} Order & Variational \\
  \cmidrule(lr){3-4} \cmidrule(lr){5-7}
Textured &$-\mathcal{F}$ \tiny{$(\times 10^4)$} &
${3.1 \,  [3.0-3.2]}$ & ${2.3 \, [2.2-2.4]}$ &
${0.90 \, [0.76-0.97]}$ & ${0.91 \, [0.77-0.98]}$ & $\mathbf{0.74 \, [0.59-1.2 \hphantom{0}]}$  \\ 
Bouncing Ball & nMSE &
${\ge 0.99}$ & ${{\ge 0.99}}$ &
${0.85 \, [0.06-1.00]}$ & ${0.14 \, [0.04-0.94]}$ & $\mathbf{0.05 \, [0.03-0.26]}$ \\ 
\cmidrule(lr){3-4} \cmidrule(lr){5-7}
Structured &$-\mathcal{F}$ \tiny{$(\times 10^3)$}&
${40 \, [28 - 46]}$ & ${26 \, [21 - 30]}$ &
${1.1 \hphantom{0}\, [1.1 \hphantom{0}- 1.1 \hphantom{0}]}$ & ${0.96 \, [0.93 - 1.0 \hphantom{0}]}$ & $\mathbf{0.77 \, [0.75 - 0.80]}$ \\ 
Background & nMSE &
$ {\ge 0.99} $ & ${\ge 0.9}$ &
${0.11 \, [0.10 - 0.14]}$ & ${0.11 \, [0.10 - 0.12]}$ & $\mathbf{0.09 [0.09 - 0.10]}$ \\ 
\end{tabular}
\end{center}
\caption{Performance on the Textured and Structured Background Bouncing Ball Experiments using negative free energy ($-\mathcal{F}$; lower better) and normalised mean squared regression error to the true latent (nMSE; lower better). We compare  different fitting procedures for RP-GPFA with a single latent dimension. sGP-VAE is fitted with both one or two latent dimensions. Values indicate median and inter-quartile range over 20 random seeds.}
\label{tab:BallBaseline}
\end{table*}

\subsubsection{Textured Bouncing Ball}
\label{sec:textBB}
\begin{figure}
\centering
\includegraphics[width= 0.9 \linewidth]{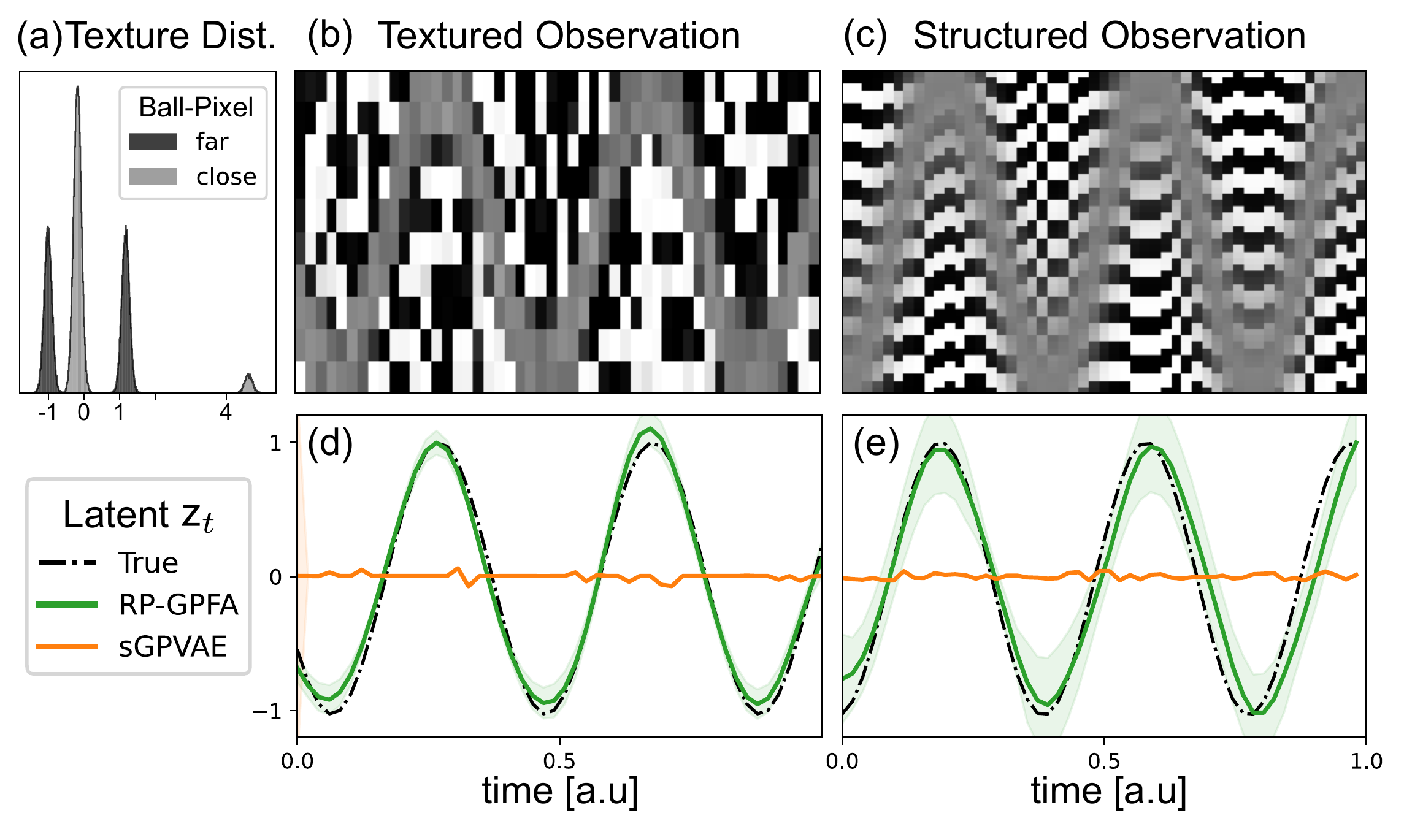}
\caption{\label{fig:textured} Bouncing ball experiments: a latent variable $\z_t$ modulates pixel intensity. (a,b) $\z_t$ influences the higher order statistics of the image (i.e., the texture). (c) Intensity modulation is imposed over structured background. (d,e) Latent recovery using RP-GPFA (variational bound method) or sGP-VAE. Shades indicates 2 standard deviations. }
\end{figure}

We illustrate RP-GPFA on a modified version of the bouncing ball experiment \citep{johnson+al:2016:svae},  in which a one-dimensional latent modulates the intensity of observed pixels across time (\cref{fig:textured}). The stochastic mapping from latent to observation is defined such that the mean and variance of pixel intensity is independent of the latent position. We compare our approach to sparse Gaussian process VAE (sGP-VAE) \citep{ashman2020sparse} and report the negative free energy ($-\mathcal{F}$) 
and the normalised mean squared error (nMSE) obtained by linear regression from inferred to true latent (\cref{tab:BallBaseline}). RP-GPFA is fit using a one dimensional latent space with each of the E-step methods described in \cref{sec:ml:estep}.
Reparametrisation employed only 20 samples to maintain computational comparability. 
All methods shared the same recognition network structure (two fully connected layers of size 50, ReLU activation function) and were trained using Adam. \cite{kingma2014adam}.

The latent variable influences the higher order statistics of the image (i.e., the texture) but the standard sGP-VAE generative model maps latent to observations through  multivariate Gaussian distributions. As a consequence, the one-dimensional version of this model is predictably blind to the latent oscillations. Interestingly, this was still the case when using a two-dimensional latent space. In contrast, the implicit generative process of RP-GPFA is not subject to model mismatch and recovers the latent dynamics accurately. The best performance was reached using the interior variational bound, albeit with high variability.  The second order approximation yielded competitive and more reliable results across the 20 random seeds.

\subsubsection{Structured-Background Bouncing Ball}
In a second variant of the bouncing ball experiment, the ball appeared as a local Gaussian blur imposed over a structured, striped, moving background (\cref{fig:textured}).  
This example helps to illustrate another shortcoming of explicit generation, beyond the risk of model mismatch illustrated above.  
The generative likelihood depends on the capacity to reconstruct the entire observation, including any structured but independent features that cannot be ascribed to noise.  
In this example, the sGP-VAE must work to model both structured background and ball-related features, which proves impossible with one or two latent processes.  
By contrast, the RPM likelihood focuses on latent structure which renders observations conditionally independent in time (\cref{fig:textured}).  
The difference is again reflected quantitatively in the free energies achieved and match between the recovered latent and ball position (\cref{tab:BallBaseline}). 

\begin{figure*}
\centering
\input{figures/video_distance_fig.tikz}
\caption{\label{fig:multiplefactors} Multi-factorial integration across time. RP-GPFA combines image frames $\x_{1t}$ and noisy range-finding data $\x_{2t}$ tracking an agent moving amongst distractors. No structural priors on the source of the data streams are included.  By identifying a 2D signal that renders the complex data streams conditionally independent, it recovers the position $\z_t$ of the range-finder-equipped agent. In particular the noisy range-finder data is related to only one of the many agents in the environment, which the RPM-reconstructed $\z_t$ learns to select.}
\end{figure*}
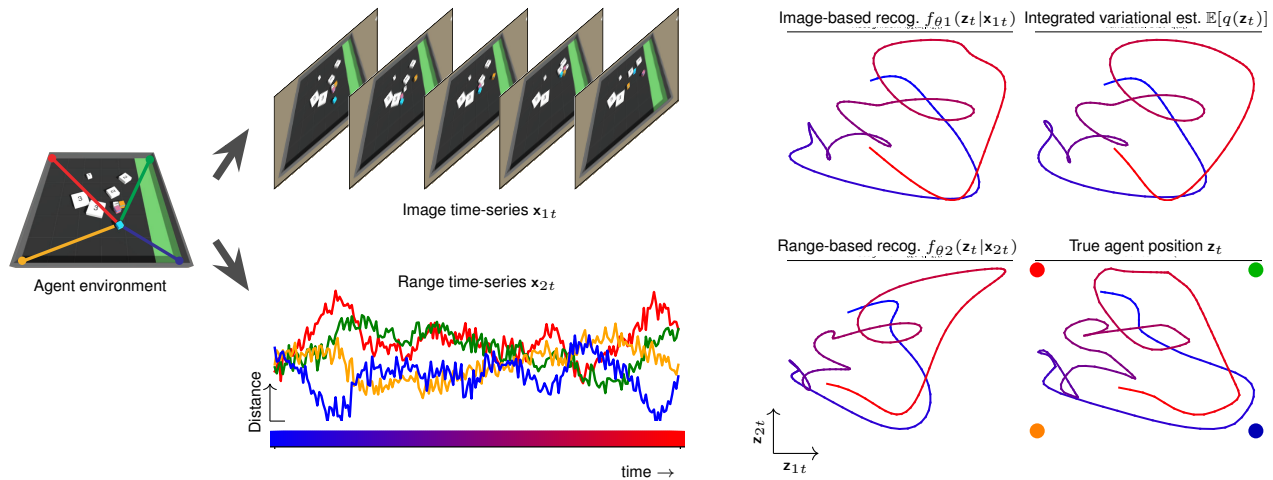

\subsubsection{Multi-factorial Integration across Time}
Last, we considered conditional independence structured across time and observed signal.  Three independent agent navigate in a bounded environment, moving inanimate blocks to a designated target.  Observations of the agents' locations are collected in the form of 3D-rendered image frames of the entire environment
 $\x_1 = \{\x_{1t}: t= 1\dots T \}$, as well as noisy range-finding sensor data giving the distances of one of the agents from the four corners of the room $\x_2 = \{ \x_{2t}: t= 1\dots T \}$ (Fig. \ref{fig:multiplefactors}).  Renderings and trajectories were generated using Unity Machine Learning Agents Toolkit \citep{juliani2018unity}. 
In this setting, a (2D) latent inducing conditional independence should recover the location of the range-finder-equipped agent, ignoring the other agents and the rest of the image data.

We trained RP-GPFA using the 2\textsuperscript{nd}-order approximation method, 40 inducing points, a convolutional network $\theta_1$ acting on image data and a two-layer perceptron $\theta_2$ on range data (resp.~similar to \cref{sec:peer} and \cref{sec:textBB}).

\cref{fig:multiplefactors} shows one full trajectory $\setZ = \{ \z_{t}: t= 1\dots T \}$, the recovered mean of the variational distribution $q$, and the individual video and range recording factors $\pthetazxj(\cdot | \x_{jt})$ combined with $\pthetaz$.
These results illustrate how the pursuit of conditional independence underlying RP-GPFA makes it possible to (i) learn the nonlinear mapping from distance to position signal and (ii) learn to track a moving agent from video recordings. Perhaps more importantly, it is the conditional independence structure across distance sensors and video that provides the signal guiding the video network to (iii) learn which agent to track amongst the distractors.




\section{CONCLUSION}
We have introduced the \textit{recognition-parametrised model}, a normalised semi-parametric family in which the latent variables model the joint dependence of observations but not their individual marginals.
As the parametric part of the likelihood is defined in terms of the recognition parameters alone, the RPM avoids issues of mismatch between generative and recognition models, and enables rapid computations of latent posterior distributions from observed data.
Furthermore, by incorporating the empirical marginal distribution of individual latents, the RPM is able to capture joint structure, regardless of details of the noise distribution, or of unrelated distractors.  

The RPM may be defined using simple exponential family forms on the latents, allowing access to the wide range of probabilistic tools. 
The capacity for structured probabilistic inference was exploited in the experiments here, with RP-LDA exemplifying the use of hierarchical models and variational Bayes, while RP-GPFA combined RPM inference with the sparse variational GP approximation.
The model can be learned through maximum-likelihood exactly in the case of discrete latents and we present several approximations to a variational bound for the continuous case. 

Animals and artificial agents acting in the world need to learn structure in sensory input to build representations of their environments and infer state, but they rarely need to generate synthetic observations. 
The assumptions of the RPM: that recognition is probabilistic, detailed simulation is avoided, and learning is unsupervised are likely to be those that shape natural intelligence.
Behavioural studies reveal Bayesian perception and decision making 
 under noise, uncertainty and risk; dense cortico-fugal connections do not extend to the sensory periphery; and natural human
``supervision'' in fact corresponds to the RPM principle: object category is the thing that makes the
utterance of a caregiver conditionally independent of the picture or object to which they are pointing.  
We are unaware of other learning frameworks that are fully probabilistic, unsupervised,   tractable, and avoid explicit instantiation of a generative model.  
Thus models of the RPM type may be central to replicating general animal-like intelligence.
and so we believe it holds promise both as a model of biological learning and as a basis for efficient state discovery and action learning in artificial settings.

\subsubsection*{Acknowledgements}
This work was funded by the Gatsby Charitable Foundation and Simons Foundation (SCGB 543039).  We thank Ted Moskovitz, Marcel Nonnenmacher and Peter Orbanz for helpful discussions.



\bibliographystyle{abbrvnat}
\bibliography{journalsabbrvnodots,rpm}

\appendix
\onecolumn

\include{rpm_aistats2023_final_supplement}

\end{document}

%% file: figures/video_distance_fig.tikz
\newcommand{\fsep}{1}
\newcommand{\includeframe}[2][]{
    \includegraphics[bb=55 55 420 285,clip,#1]{figures/video_and_distance/observation#2.pdf}
}
\newcommand{\includerecog}[2][]{
    \includegraphics[bb=70 50 410 310,clip,width=3cm,#1]{figures/video_and_distance_2/#2.pdf}
}

\begin{tikzpicture}[
    every node/.style={font={\sf\tiny}},
    img/.style={inner sep=0pt}
    ]

  \node (frame0) at (0,0) [img,left]
   {\includegraphics[width=2.5cm]{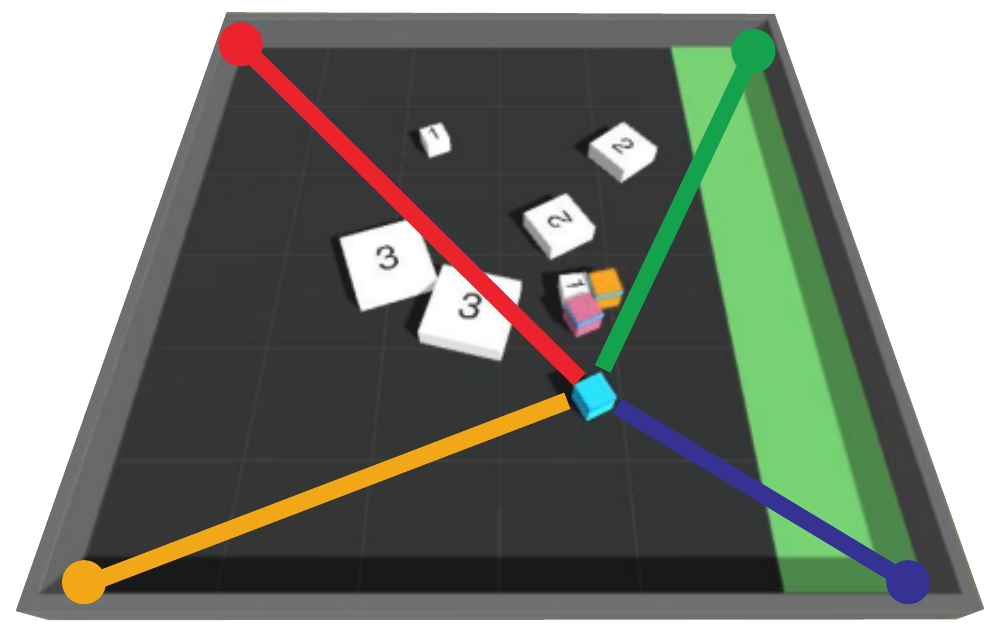}};
  \node [below=0mm of frame0] {Agent environment};

  \foreach \ff in {1,...,5} {
    \node (frame\ff) [img, 
        transform canvas={shift={(\ff*\fsep,0.2)},xscale=0.7,yslant=0.6},
        above right]
        {\includeframe[width=2cm]{\ff}};
  }
  \node at (3.8*\fsep,0.2) [below] {Image time-series $\x_{1t}$};

  \node (range) at (\fsep,-1) [img,below right]
     {\includegraphics[width=5.5cm,bb=55 40 315 125,clip]{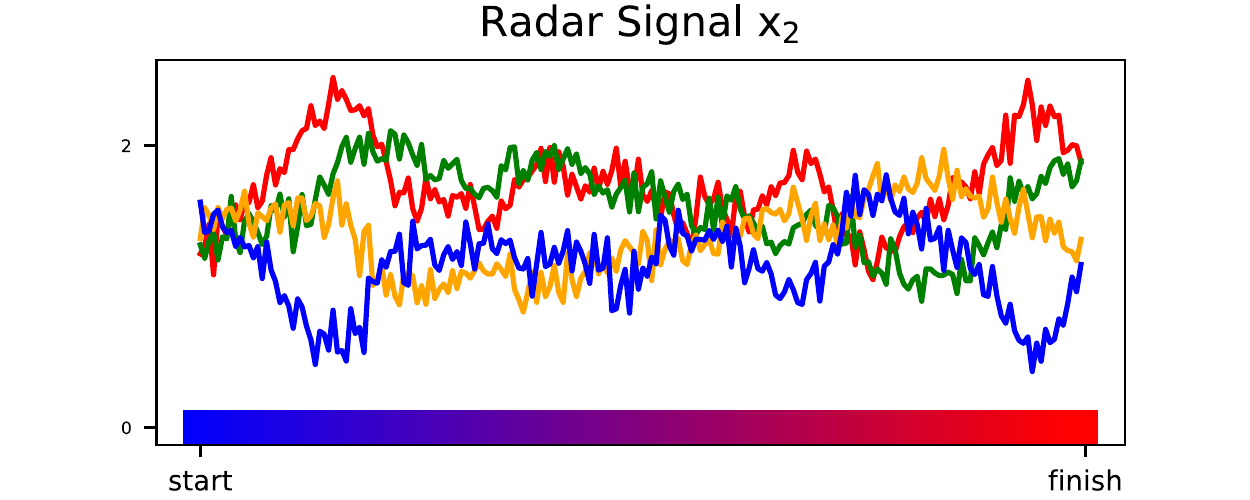}};

  \node [above=-2mm of range] {Range time-series $\x_{2t}$};
  \draw[->] (range.south west) -- +(0,0.5) node [midway,above,rotate=90] {Distance};
  \draw (range.south west) -- +(0.2,0);

  \node (time) [img,below=1mm of range]
     {\includegraphics[width=5.5cm,bb=55 15 315 27,clip]{figures/video_and_distance_2/distance_signal.pdf}};

  \node [below left=1mm and 0mm of time.south east] {time $\to$};

  \draw[-Stealth,line width=1mm,shorten <=5mm, shorten >=5mm,black!70] (frame0.east) -- (\fsep,1.5);
  \draw[-Stealth,line width=1mm,shorten <=5mm, shorten >=5mm,black!70] (frame0.east) -- (\fsep,-1.5);

  \begin{scope}[shift={(11,0)}]
    \node (recog_f1)   [above left] {\includerecog{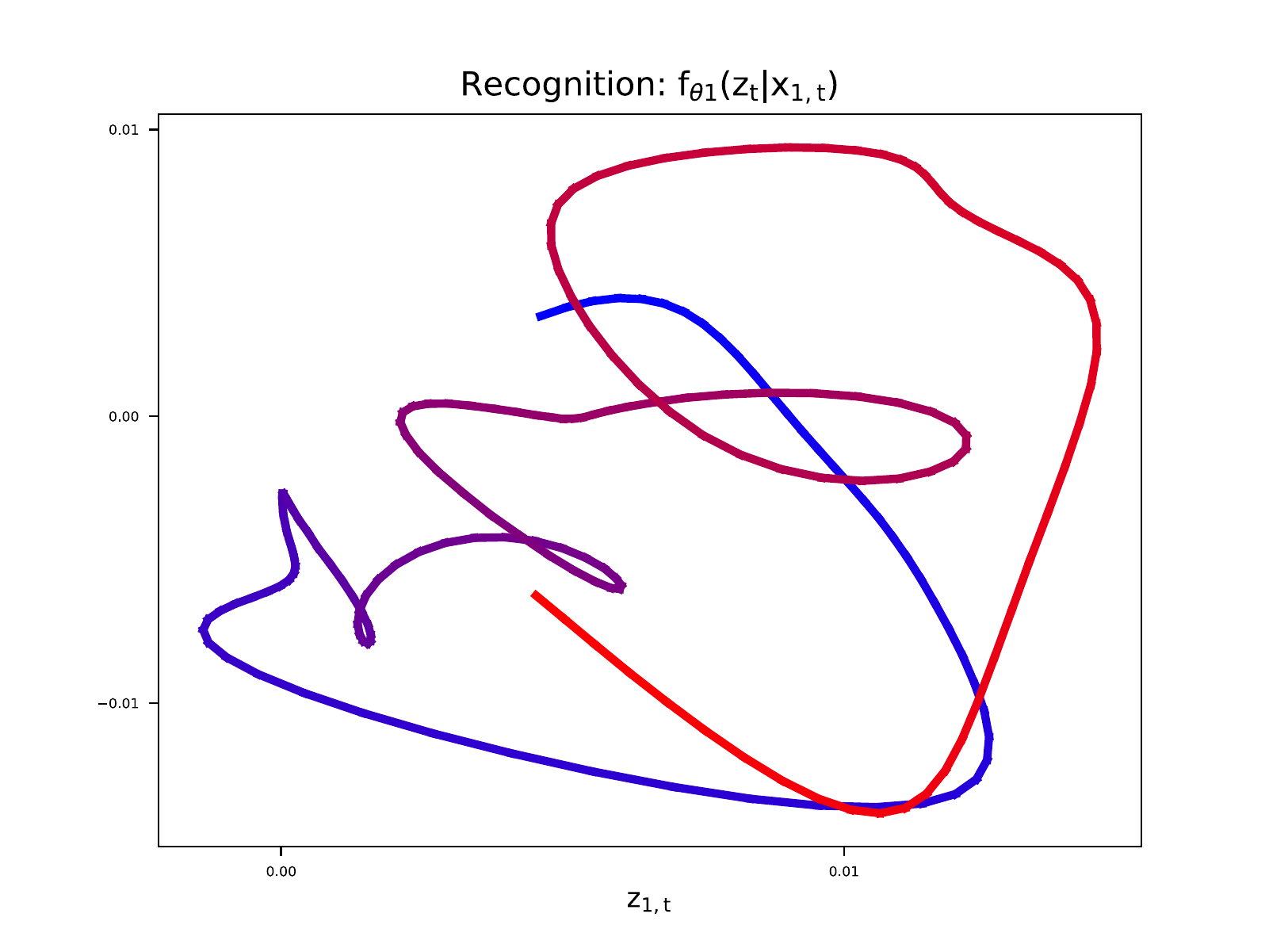}};
    \node (recog_f2)   at (0,-0.5) [below left]{\includerecog{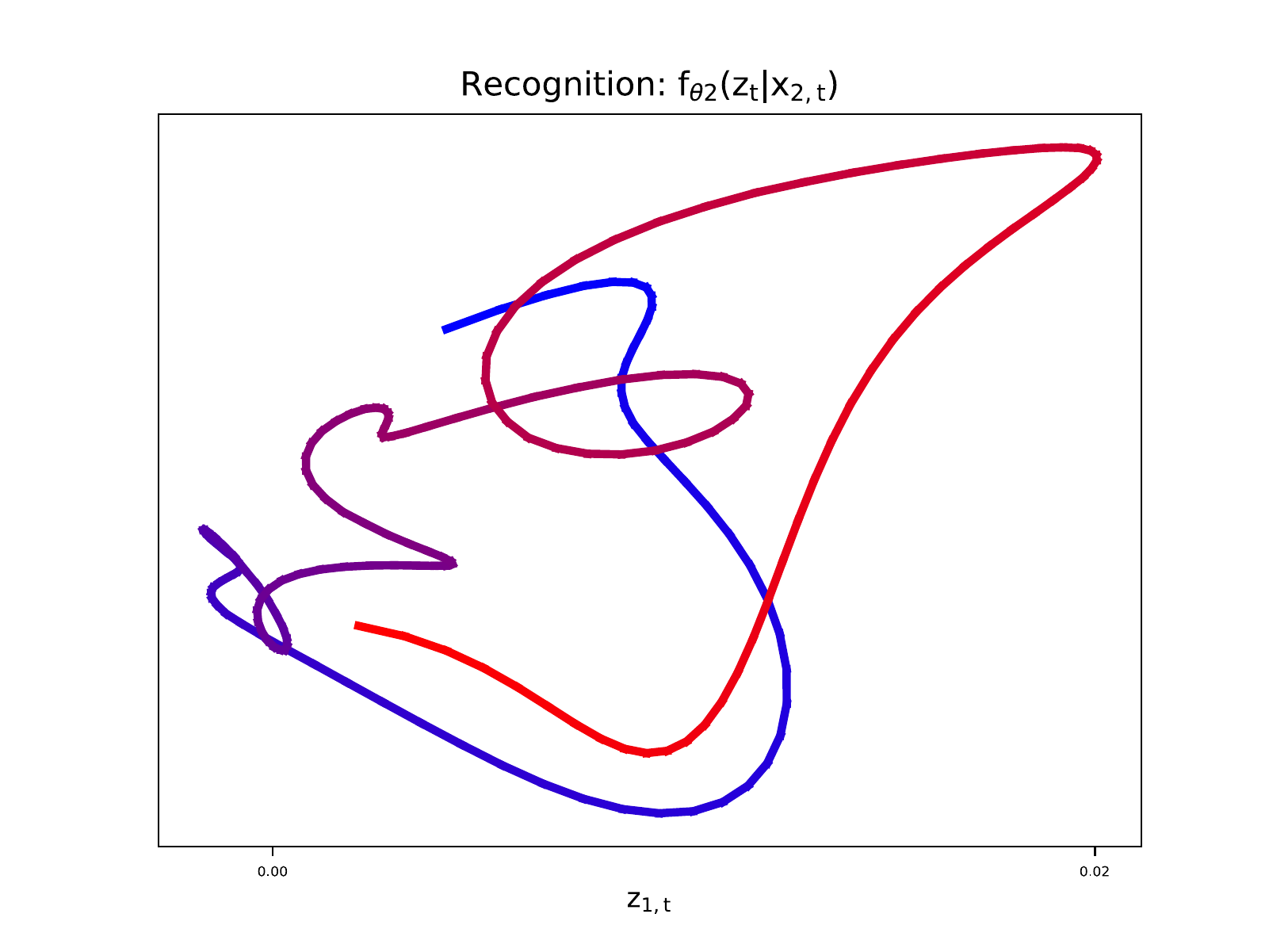}};
    \node (recog_var)  [above right]{\includerecog{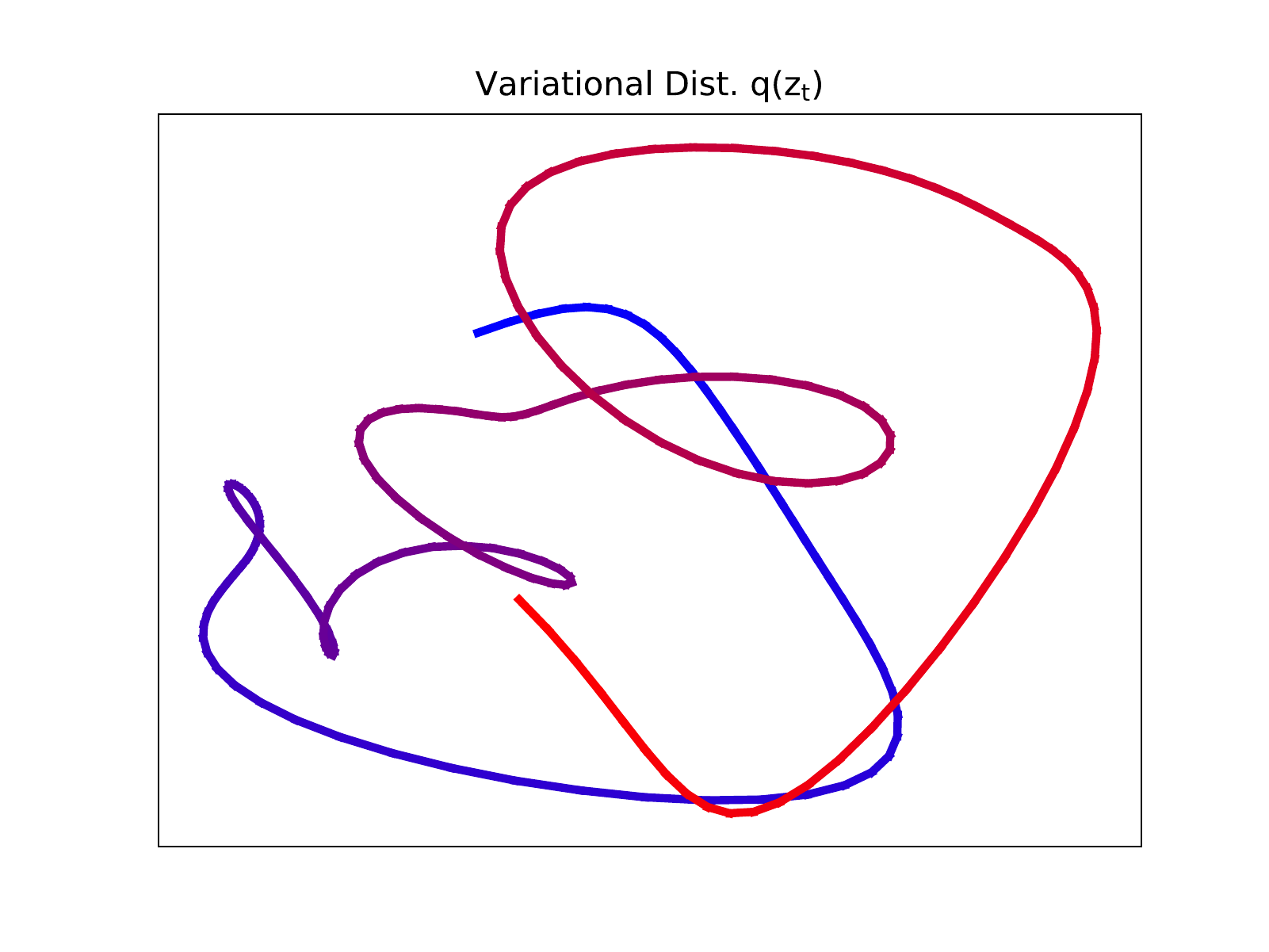}};
    \node (recog_true) at (0,-0.5) [below right] {\includerecog{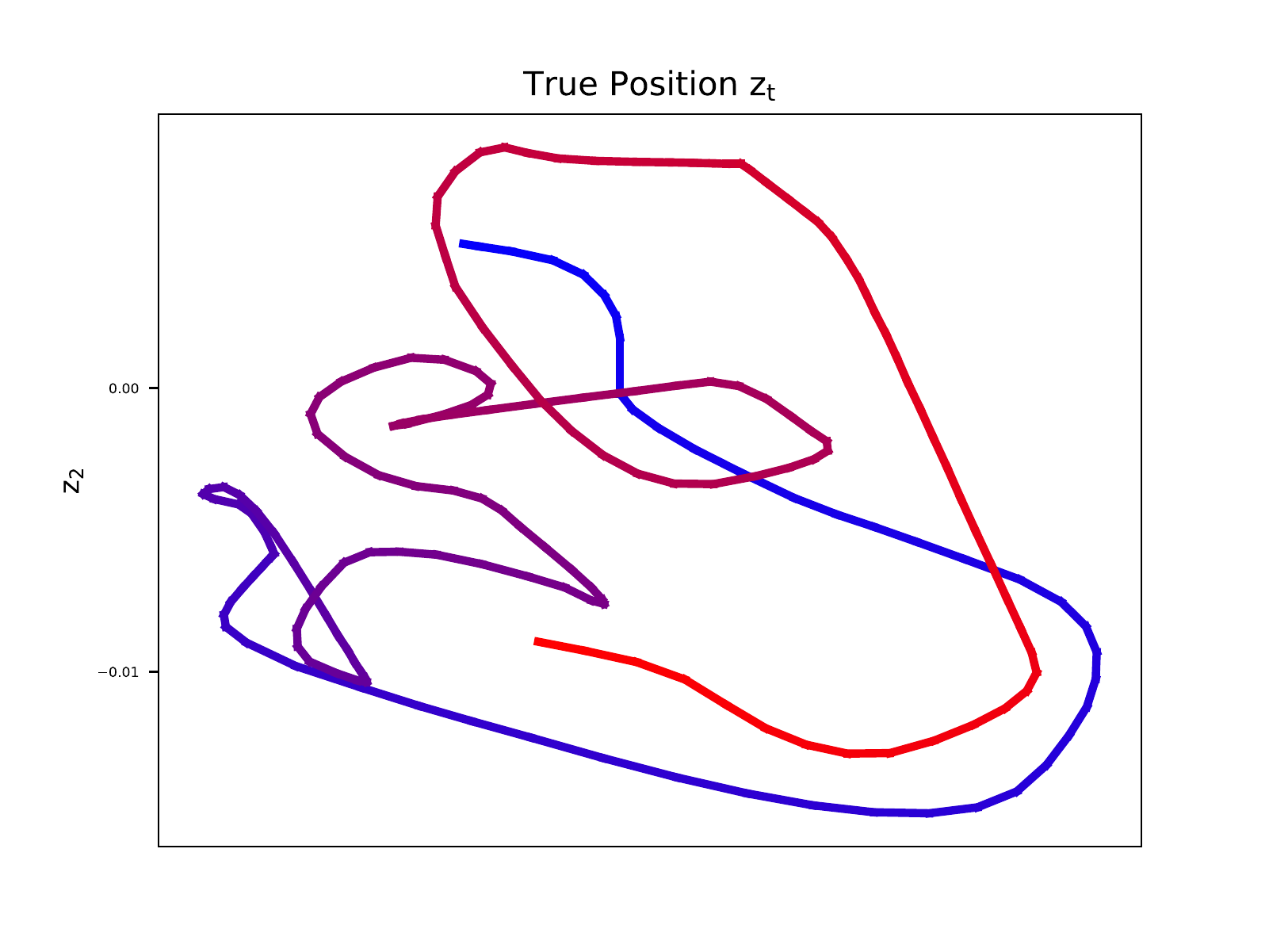}};

    \fill[red] (recog_true.north west) ++(0.2,-0.3) circle[radius=1mm];
    \fill[green!70!black] (recog_true.north east) ++(-0.2,-0.3) circle[radius=1mm];
    \fill[orange] (recog_true.south west) ++(0.2,0.1) circle[radius=1mm];
    \fill[blue!70!black] (recog_true.south east) ++(-0.2,0.1) circle[radius=1mm];

    \node at (recog_true.north) {True agent position $\z_t$};
    \node at (recog_var.north) {Integrated variational est.\ $\expect{q(\z_t)}$};
    \node at (recog_f1.north) {Image-based recog.\ $\pthetazxj[1](\z_t|\x_{1t})$};
    \node at (recog_f2.north) {Range-based recog.\ $\pthetazxj[2](\z_t|\x_{2t})$};

    \draw[->] (recog_f2.south west) ++(0,-0.2) -- +(0.6,0) node[midway,below] {$\z_{1t}$};
    \draw[->] (recog_f2.south west) ++(0,-0.2) -- +(0,0.6) node[midway,above,rotate=90] {$\z_{2t}$};

  \end{scope}
\end{tikzpicture}

%% file: rpm_aistats2023_final_supplement.tex

\def\E{\mathbb{E}}
\def\V{\mathbb{V}}
\def\m{\vec{m}}

%

%

\onecolumn



\setcounter{figure}{0}\renewcommand{\thefigure}{A\arabic{figure}}
\setcounter{table}{0}\renewcommand{\thetable}{A\arabic{table}}
\setcounter{equation}{0}\renewcommand{\theequation}{A\arabic{equation}}
\newtheorem{prop}{Proposition}


\section{MAXIMUM LIKELIHOOD LEARNING}
\label{appendix:ml-learning}

We provide further details and complete derivations of key results.  Equation numbers without an 'A' prefix correspond to those in the main text.

Recall that the full joint distribution associated with a Recognition-Parametrised Model (RPM) takes the form
\begin{equation}\tag{2}  
   \pthetaXN{\setX, \setZ}= \pthetaz(\setZ) 
     \prod_j \lr[\Big](){
        \pxempj(\x_j)
        \frac{\pthetazxj(\setZ| \x_j)}{\pthetazj(\setZ)}
    }  \,,
\end{equation}
where $\setX = \{\x_j : j = 1\dots
J\}$ is a set random variables and $\setZ = \{\z_l : l =
1\dots L\}$ is a set of underlying latent variables given which $\x_j$ are conditionally independent. The factors are defined as
\begin{description}[labelindent=0em]
  \item[$\pthetaz(\setZ)$]: a normalised distribution on the latent variables whose factorisation depends on a latent graphical model. 
  \item[$\pxempj(\x_j) = \frac 1N\sum_n \delta(\x_j - \x_j\nn)$]: the empirical measures with atoms at the $N$ data points $\x_j\nn$. 
  \item[$\pthetazxj(\setZ| \x_j)$]: parameterised distributions that we call ``recognition factors''.
  \item[$\pthetazj(\setZ) = \int d\x_j \pxempj(\x_j) \pthetazxj(\setZ | \x_j)$]: mixture of recognition factors with respect to the empirical measures.
\end{description}
As we take $\pxempj(\x_j)$ to be the empirical measures throughout, the mixtures have the forms $\pthetazj(\setZ)  = \frac1N \sum_n \pthetazxj(\setZ| \x_j\nn)$.

\subsection{Variational Free Energy}
We use Expectation-Maximisation coordinate ascent of the variational free energy (sometimes referred to as Evidence Lower Bound or ELBO) derived by applying Jensen's inequality to the log likelihood \citep{neal+hinton:1998}:
\begin{align}
    \lefteqn{\sum_n \log \pthetaXN{\setX\nn} \ge
    \eff(\theta, q(\{\setZ\nn\})) 
    }\quad&
    \nonumber \\
    &= \angles[\Big]{\sum_n\log\pthetaXN{\setX\nn,\setZ\nn}}
      + \entropy{q}
    \nonumber \\
    &= \sum_n \lr[\Bigg](){
      \angles[\big]{\log \pthetaz(\setZ\nn)}
      + \sum_j \lr[\Big](){
        \angles[\big]{\log \pthetazxj(\setZ\nn | \x_j\nn)}
      - \angles[\big]{\log  \pthetazj(\setZ\nn) }
      }} \nonumber
    \\
    &\qquad + \sum_{jn} \log \pxempj(\x_j\nn) + \sum_n \entropy[\big]{q\nn} 
    \label{eq:FullFE}
\end{align}    
where, as in the main text, angle brackets indicate expectations with respect to the variational distribution $q$, $\entropy{\cdot}$ is the entropy, and we have used the fact that the optimal $q$ has the form $\prod q\nn(\setZ\nn)$. 
When the latent variables that appear in $\pthetazj(\setZ)$ are discrete, and the graphical structure of $\pthetaz$ admits exact belief propagation, the expressions in \cref{eq:FullFE} can be evaluated in closed form and optimisation is straightforward.  For the more challenging case of continuous-valued latent variables, we introduced three approximation approaches in the main text.  These are reviewed and developed further below.  

\subsection{E-step for Continuous-Valued Latent Variables}
We consider the case in which  $\pthetaz$, $\pthetazxj(.| \x_j)$ and $q\nn$ are all members of the same exponential family with natural parameters $\etaz$, $\etaxjn$ and $\etaqn$ respectively, corresponding to minimal sufficient statistic $t(\setZ)$ and log-normaliser $\Phi$. The expectations of $t(\setZ)$ under the corresponding distribution are written $\muz$, $\muxjn$ and $\muqn$ respectively (notation is summarised in table \ref{tab:tnotations}). With this set of assumptions, the only term from
\cref{eq:FullFE} that cannot be expressed analytically is $\angles[\big]{\log  \pthetazj(\setZ\nn)}$. As discussed in the main paper, if a sample from $q$ can be expressed as a parametrised function of a sample from a fixed distribution (as is the case with multivariate normal distributions) it can be evaluated using Monte-Carlo estimates. Nevertheless, this approach may be computationally expensive for high dimensional problems. We therefore propose two additional approaches to handle intractable terms.
%
\begin{table}
\begin{center}
\def\NA{\textcolor{black!30}{\small N/A}}
\begin{tabular}{|c|c|c|c|c|c|} 
 \hline
 Name & Distribution & Natural Parameter & $\E(t(\setZ))$ & $\V(t(\setZ))$ & Normalised ? \\ 
 \hline
 Prior &$\pthetaz$ & $\etaz$ & $\muz$ &  \NA & Yes \\ 
 Recognition Factors&$\pthetazxj(.| \x_j)$ & $\etaxjn$ & $\muxjn$ &  \NA & Yes\\ 
 Variational &$q\nn$ & $\etaqn$ & $\muqn$ & $V_q\nn$ & Yes\\ 
 Auxiliary Factors &$\tilde f_j\nn$ & $\tilde \eta_j\nn$ & \NA &  \NA & No\\ 
 Normalised Auxiliary &$\hat f_j\nn$ & $\etaxjn - \tilde \eta_j\nn$ &  \NA &  \NA & Yes\\ 
 Mixture &$\pthetazj$ &  \NA &  \NA &  \NA & Yes\\ 
 \hline
\end{tabular}
\end{center}
 \caption{ \label{tab:tnotations} Notation Glossary for Continuous Exponential Family Case }
\end{table}
\subsubsection{Second-order Approximation}

First, we generalise the approach introduced by \citep{braun2010variational} and expand $g(t(\setZ)) = \log \pthetazj(\setZ)$ to second order in $t(\setZ)$ around its expectation $\muqn$. This gives
\begin{equation}
    g(t(\setZ)) \approx g(\muqn) 
    + \partial g^\top \lr(){t(\setZ) - \muqn}
    + \frac12 \lr(){t(\setZ) - \muqn}^\top \partial^2 g\, \lr(){t(\setZ) - \muqn}\,,
\end{equation}
where
\begin{equation}
    \partial g = \sum_m \eta(\x_j\nn[m]) \pi_{jm}\nn
    \,,
\end{equation}
\begin{equation}
    \partial^2 g = \sum_m \eta(\x_j\nn[m]) \eta(\x_j\nn[m])^\top \pi_{jm}\nn
    -
    \sum_{m,m'} \eta(\x_j\nn[m]) \eta(\x_j\nn[m'])^\top \pi_{jm}\nn \pi_{jm'}\nn
    \,,
\end{equation}
with
\begin{equation} \tag{7}
    \pi_{jm}\nn = \frac{e^{ \eta(\x_j\nn[m])^\top \muqn  - \Phi(\eta(\x_j\nn[m]))}}{\sum_{p} e^{ \eta(\x_j\nn[p])^\top \muqn - \Phi(\eta(\x_j\nn[p]))}} \,.
\end{equation}
The first order term vanishes when taking the expectation over $q\nn$ so that
\begin{equation}
    \langle g(t(\setZ)) \rangle
    \approx g(\muqn) 
    + \frac12
    \text{tr} 
    \lr(){
    V_q\nn
     \partial^2 g} \,.
\end{equation}
Finally, we gather the recognition factor natural parameters in 
\begin{equation*}
    \vec{\eta}_j = \left[
    \eta(\x_j^{(1)}), \dots, \eta(\x_j^{(N)}) \right] 
\end{equation*}
and the weights in
\begin{equation*}
    \vec{\pi}_j\nn = 
    \left[
    \pi_{1j}\nn, \dots, \pi_{Nj}\nn
    \right] ^\top \,,
\end{equation*}
yielding
\begin{equation} \tag{6}
\begin{split}
    \big\langle \log \pthetazj \big\rangle &\approx 
    \log \frac{1}{N} \sum_{m=1}^N e^{ \eta(\x_j\nn[m])^\top \muqn - \Phi(\eta(\x_j\nn[m]))} 
    + \frac12  \text{tr}
    \left(
    \vec{\eta}_j^\top V_q\nn \vec{\eta}_j \left[
    \text{diag}(\vec{\pi}_j\nn) - \vec{\pi}_j\nn{\vec{\pi}_j\nn}^\top
    \right]
    \right)\,.
    \end{split}
\end{equation}
This form can be inserted into \cref{eq:FullFE} to yield a tractable, approximate free energy. 

In the case where $q\nn$ is a multivariate distribution with mean $\m \nn$ and variance $S\nn$, we recall 
\begin{equation}
    \muqn = 
    \left[
\begin{matrix}
\m \nn\\
\text{Vec}\left( S\nn + \m \nn {\m \nn}^\top \right)
\end{matrix}
\right] 
\end{equation}
and
\begin{equation}
    V_q\nn = 
    \left[
\begin{matrix}
S\nn & \m \nn {}^\top \otimes S\nn + S\nn \otimes \m  \nn {}^\top\\
\m \nn \otimes S\nn + S\nn \otimes \m \nn & \mathbb{S}\nn
\end{matrix}
\right] 
\end{equation}
where 
\begin{equation*}
    \mathbb{S}\nn =
    h(S\nn, S\nn) + 
    h(S\nn, \m \nn \m \nn {}^\top) +
    h(\m \nn \m \nn {}^\top, S\nn) \,.
\end{equation*}

and

\begin{equation}
    h(A, B) =
    A \otimes B
    +
    \left( \Gamma^\top
    \otimes 
    A
    \otimes 
    \Gamma
    \right)
    \odot
    \left( \Gamma
    \otimes 
    B
    \otimes 
    \Gamma^\top
    \right) \text{ with } \Gamma = 1_{K \times 1}
\end{equation}

$\otimes$ and $\odot$ are the Kronecker and Hadamard products. \\

\subsubsection{Interior Variational Bound}

The previous approach gives a compact approximation of the free energy but it is not guaranteed to lower bound the log-likelihood. Thus, we considered a second strategy in which we introduced a further relaxation of the free-energy bound, by introducing auxiliary functions $\tilde f_j\nn(\setZ)$. Focusing on the
$\pthetazj$-dependent terms as above, we have
\begin{equation}
\begin{split}
\angles{\log \frac{\pthetazxj(\cdot| \x_j\nn)}{\pthetazj}}
    &= \angles{\log \frac{\pthetazxj(\cdot| \x_j\nn)}{\tilde f_j\nn q\nn}}
          - \angles{\log \frac{\pthetazj}{\tilde f_j\nn q\nn}}
  \\
    &\ge \angles{\log \frac{\pthetazxj(\cdot| \x_j\nn)}{\tilde f_j\nn q\nn}}
          - \log \angles{\frac{\pthetazj}{\tilde f_j\nn q\nn}}
    \quad \text{(by Jensen)} 
  \\
    &= \angles{\log \frac{\pthetazxj(\cdot| \x_j\nn)}{\tilde f_j\nn q\nn}}
          - \log \intdx[d\setZ] \frac{\pthetazj(\setZ)}{\tilde f_j\nn(\setZ)} \,.
    \label{eq:dvsup}
\end{split}
\end{equation}
If we now choose $\tilde f_j\nn(\setZ) = \exp\lr[\big](){t(\setZ)\tr \tletajn}$ with the constraint that $\etaxjn[m] - \tletajn$ is a valid natural parameter for all $(m, n)$, then the right hand-side of \cref{eq:dvsup} is closed-form
\begin{equation}
\begin{split}
    \tilde \Gamma_j \nn = \int d \setZ \frac{\pthetazj(\setZ)}{\tilde f_j\nn(\setZ)}
    = \frac1N \sum_m e^{\Phi(\eta_j(\x_j\nn[m]) - \tilde\eta_j\nn) - \Phi_j(\eta_j(\x_j\nn[m]))} \,.
  \end{split}
\end{equation}
By rearranging terms, we obtain (c.f.\ main text eq.~8)
\begin{equation}
\begin{split}
\bigg\langle \log \frac{\pthetazxj(\z_j | \x_j\nn)}{\pthetazj(\z_j)} \bigg\rangle \geq &
- \KL{q\nn}{ \hat{f}_j\nn } + \log \Gamma_j\nn \,,
\end{split}
\end{equation}
where $\hat{f}_j\nn$ is a properly normalised exponential family distribution with natural parameter $\eta_j(\x_j\nn) - \tilde\eta_j\nn$ and
\begin{equation}
    \Gamma_j\nn = 
    \frac
    {e^{\Phi(\eta_j(\x_j\nn[n]) - \tilde\eta_j\nn) -\Phi(\eta_j(\x_j\nn[n]))}}
    {\frac1N \sum_m e^{\Phi(\eta_j(\x_j\nn[m]) - \tilde\eta_j\nn) - \Phi_j(\eta_t(\x_j\nn[m]))}} = 
    \frac
    {e^{\Phi(\eta_j(\x_j\nn[n]) - \tilde\eta_j\nn) -\Phi(\eta_j(\x_j\nn[n]))}}
    {\tilde \Gamma _j \nn} \,.
\end{equation}
This fully tractable expression can then be inserted in \cref{eq:FullFE}. Furthermore, the terms of the resulting expression can be rearranged to make explicit the bound to the conventional free energy and the log-likelihood
\begin{equation}
\begin{split}
\sum_n \log\pr[\theta,\XN][\big]{\setX\nn} \geq \eff(\theta,q) &\geq \widetilde\eff\lr(){\theta, q, \{\tilde f_j\nn\}} \,,
\end{split}
\end{equation}
where
\begin{equation} \tag{9}
\begin{split}
 \widetilde\eff\lr(){\theta, q, \{\tilde f_j\nn\}} = \sum_n \log\pr[\theta,\XN][\big]{\setX\nn} 
    - \sum_n \KL[\big]{q\nn}{\pr[\theta,\XN][\big]{\!\cdot\!|\setX\nn}}
    - \sum_{nj} \KL[\Big]{q\nn}{\frac1{\tilde\Gamma_j\nn} \frac{\pthetazj}{\tilde f_j\nn}} \,.
\end{split}
\end{equation}
Thus, as might be expected, the second variational relaxation introduces a further KL-divergence penalty, beyond the term in $\KL{q(\setZ)}{p(\setZ|\setX)}$ introduced by the standard variational approach.

\section{DISCRETE EXPERIMENTS}
\label{appendix:exp-discrete}

In all discrete experiments, the recognition network $\theta$ was shared across factors and comprised 2 convolutional layers with max pooling and one fully connected layer of 50 units followed by a ReLU (Rectified Linear Unit) activation function.

\subsection{Peer-Supervision}\label{sec:peer_super_supplement}

In this case, the RPM had a single categorical latent variable $\z$ (so $L=1$) with uniform prior, and $J=2$ observations $\x_j$ each corresponding to an MNIST image.  The data set comprised random pairs of images of the same digit, with each digit appearing in only one pair. The factors $\pthetazxj(\z|\x_j)$ were parametrised by a single convolutional neural network (i.e., the parameters $\theta_j$ were tied), which outputs categorical probabilities.  Inference is thus conjugate. Assuming that $\z$ can take $K=10$ values, the E-step has the closed form: 
\begin{equation}
    q\nn (\z = k) \propto \prod_j \frac{f_\theta(k | \x_j\nn)}{\sum_{m} f_\theta(k | \x_j\nn[m])}\,.
\end{equation}

The RPM is compared to Gumbel Softmax Variational Autoencoder (GS-VAE) 
\citep{maddison2016concrete}, Vector Quantised Variational Autoencoder (VQ-VAE) \citep{van2017neural}\footnote{\url{https://github.com/bshall/VectorQuantizedVAE }} and Gumbel-Softmax VQ-VAE (GS-VQVAE) \citep{sonderbypoole2017}\footnote{\url{https://github.com/YongfeiYan/Gumbel_Softmax_VAE}} (temperature of 0.5) . The Gumbel-Max reparametrisation trick allows the sampling of discrete random variables to be a sum of a deterministic function of the discrete probabilities and a fixed noise distribution, followed by an argmax operation. The Gumbel Softmax replaces the argmax with a softmax operation such that the gradients of the probabilities can be calculated. Thus this allows the VAE to learn using samples of the discrete latents in the loss. The VQ-VAE is a deterministic autoencoder whose encoder produces a continuous vector that then gets compared to a nearest neighbour embedding. The nearest neighbour is then used in the decoder for reconstructing data. Gradients are passed using the straight through estimator and the encoder, decoder, and nearest neighbour embedding is learned. The GS-VQVAE computes the variational posterior using the distances from encoder output to nearest neighbour embedding vectors as logits of a categorical distribution. Then learns to maximise the free energy using the Gumbel Softmax reparametrisation trick.

All VAE models shared the same generative neural network and all methods fundamentally shared the same recognition network. They differ in that the output dimension of RPM and GS-VAE was of dimension 10, while Vector Quantised Models recognition networks first output to an embedding space of dimension 64 before being mapped to one of 10 categories.




Each model was fit 10 times with different random initialisation. Once fit, the output of the recognition factor neural network was evaluated for classification accuracy on the MNIST test dataset on the basis of the best mapping from network output to digit identity (using Kuhn–Munkres algorithm).


\cref{fig:peer-sup-confusion} shows the accuracy achieved for each random seed as a function of the entropy of the average posterior.  
The RPM (alone) achieved performance of 96.5\% for 3/10 random initialisations, but in other cases drew sharp classification boundaries that confused or divided single digit classes, as seems reasonable given the lack of label supervision.  This effect can be seen in the confusion matrices shown in \cref{fig:peer-sup-confusion}.  None of the baseline models achieved better than about 80\% accuracy, and all of them created more distributed errors, confusing examples of many digit types.

\begin{figure}
\centering
\input figures/peer_supervision/peer_supervision_confusion.tikz
\caption{Accuracy and entropy of average posterior achieved by each of the four model types initialised with 10 different random seeds.  Insets show confusion matrices (for the best digit assignment of latent values) for the least and most accurate RPM, and most accurate example of the baseline models.}
\label{fig:peer-sup-confusion}
\end{figure}

\subsection{Latent Dirichlet Allocation (LDA)}
\label{sec:supp-lda}

The goal of the RPM-LDA is to infer the statistics of local image properties in natural images.  We start by decomposing each image into sub-patches and denote:

\begin{minipage}{0.2\linewidth}
  {\small\setlength{\gmu}{5ex}
  \graphmod{graphical_models/lda}[][]}\\
  \hspace*{.8em}LDA DAG\vphantom{p} 
\end{minipage}
\begin{minipage}{0.3\linewidth}
    \begin{center}
      {\small\setlength{\gmu}{5ex}
       \graphmod{graphical_models/lda}[rpm][]}\\  
      RPM parametrisation
    \end{center}
\end{minipage}\hfil
\begin{minipage}{0.50\linewidth}\leftmargini=1em
\begin{itemize}
    \item $\x_{j}\nn$ the $j$-th patch of image $n$
    \item $\z_{j}\nn$ the categorical identity of $\x_{j}\nn$
    \item $\omega\nn$ the distribution of categories for image $n$
    \item $p(\omega) = \text{Dirichlet}(\alpha, ..., \alpha)$ the prior over the category distribution
    \item $\theta$ the recognition network shared for all the patches that outputs the probabilities that a patch $\x_{j}\nn$ belongs to each category.
\end{itemize}
\end{minipage}\\[2ex]

The RPM has the form
\begin{equation}
\hfill
   \pr[\theta,\alpha,\XN]{\setX, \setZ}= \prod_{n=1}^N \prod_{j=1}^J 
     \pxemp(\x_j\nn)
     \frac{ f_\theta(\z_j\nn| \x_j\nn)}{\frac{1}{N} \sum_{m} f_\theta(\z_j\nn[n]| \x_j\nn[m])}
      p \lr[\big](){\z_j\nn | \omega\nn} p \lr[\big](){\omega\nn | \alpha}\,,
\end{equation}
where $\setZ = \big\{  \{\z_j\nn \}_j, \omega\nn \big\}_n$. 

We model the variational distribution as
\begin{equation}
    q(\setZ) = \prod_{n=1}^N q\nn_{\omega}(\omega\nn) \prod_{j=1}^J q_j\nn(\z_j\nn)
\end{equation}
where 
\begin{equation}
    q_\omega\nn = \text {Dirichlet} (\alpha_1 \nn, ..., \alpha_K\nn) 
    \quad \text{and} \quad
    q_j\nn(\z_j\nn=k) = \gamma\nn_{jk} \,.
\end{equation}
The E-Step is closed form and follows
\begin{equation}
    \alpha_k\nn = \alpha + \sum_{j=1}^J \gamma_{jk}\nn 
    \quad \text{and} \quad 
    \gamma_{jk}\nn \propto \exp \left(
    \Psi(\alpha_k\nn)
    + \log f_\theta(k | \x_j\nn) - \log f_\theta(k)
    \right)
\end{equation}
where $\Psi$ is the digamma function. 

During the M-Step, the recognition model is updated using Adam \citep{kingma2014adam} on the free energy.
\begin{figure}
\centering
\includegraphics[width= 0.7 \linewidth, trim= 0cm 0cm 0cm 0cm, clip]{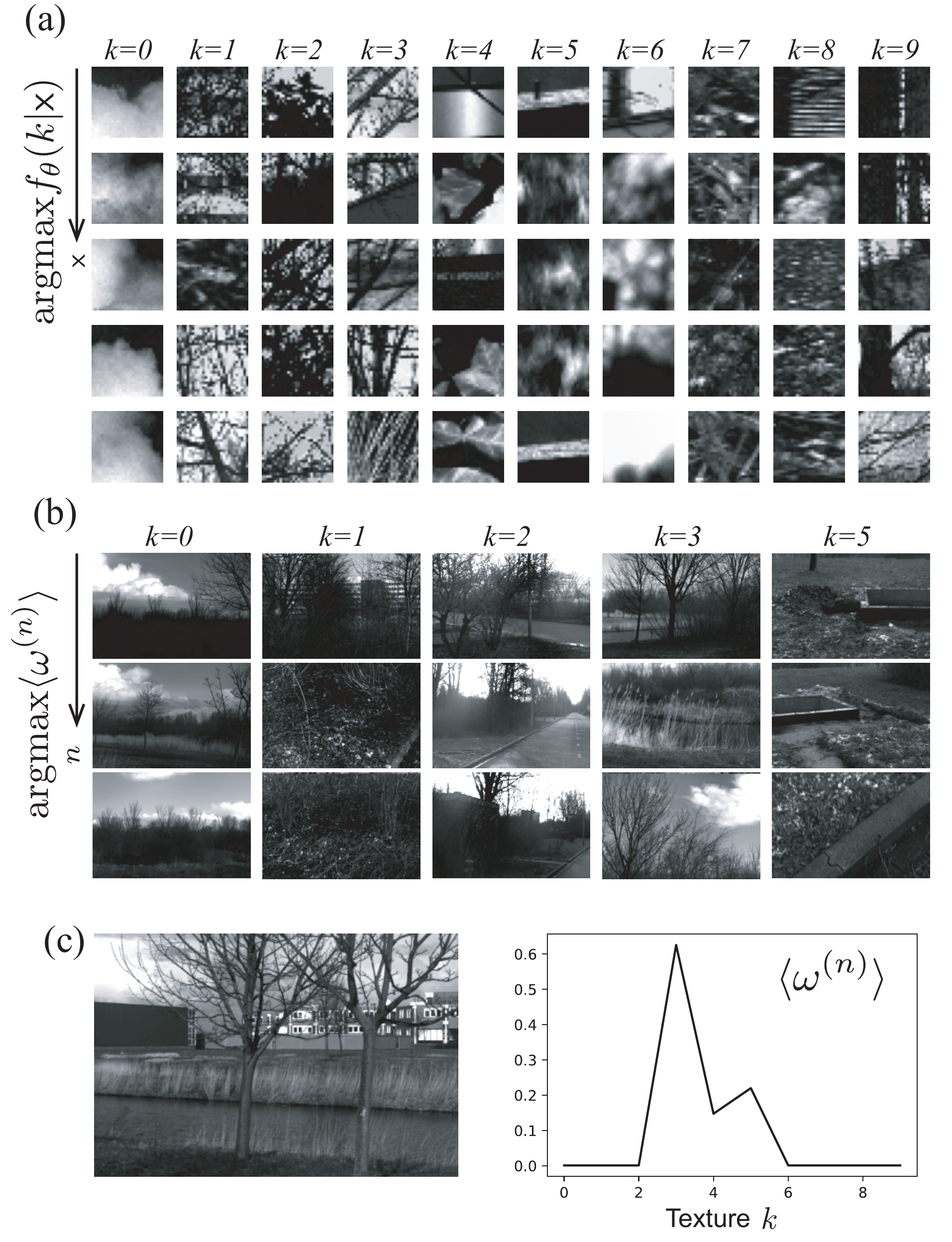} 
\caption{\label{fig:lda-supp} RPM-LDA. (a) Five most representative patches for all textures. (b) Three most representative images for some textures. (c) Texture distribution of a given image.}
\end{figure}
We applied RPM-LDA to 100 images from the van Hateren database and fixed $K=10$. Given a texture $k$, its most representative patch is the one maximising the probability of being assigned to $k$: $f_\theta(k |\x)$. We plot such patches Fig. \ref{fig:lda-supp}-(a), and see that RPM-LDA learns meaningful textural information (clouds, branches, etc.). The statistics of each image can be described by $\bar{\omega}\nn = \langle \omega \nn \rangle_q$. We confirmed the inferred textural grouping by reporting examples of images with low entropy on \ref{fig:lda-supp}-(b) and one where $\omega\nn$ is multimodal \ref{fig:lda-supp}-(c).

\section{CONTINUOUS EXPERIMENTS: RP-GPFA}

\begin{minipage}{0.48\linewidth}
\begin{center}
\vspace{0.6cm}    
  {\small\setlength{\gmu}{5ex}
   \graphmod{graphical_models/gpfa_gen}[gen][]}\\  
   \vspace{0.7cm}
  Generative Model
\end{center}
\end{minipage}
\begin{minipage}{0.48\linewidth}
\begin{center}
  {\small\setlength{\gmu}{5ex}
   \graphmod{graphical_models/gpfa_rpm}[rpm][]}\\  
  RP-GPFA parametrisation
\end{center}
\end{minipage}\label{fig:FAgenerative}\hfil
\vspace{0.5cm}

Recognition-Parametrised Gaussian Process Factor Analysis (RP-GPFA) models continuous multi-factorial temporal dependencies. We consider $J$ observed time-series measured over $T$ timesteps: $\setX = \{\x_{jt} : j = 1\dots J, t=1\dots T\}$. We seek to capture both spatial and temporal structure in a set of $K$-dimensional underlying latent time-series $\setZ = \{\z_t : t = 1\dots T\}$, such that the observations are  conditionally independent across series and across time. The full joint has the form:
\begin{equation}
   \pthetaXN{\setX, \setZ}= 
   \prod_{j=1}^J
   \prod_{t=1}^T \left(
     \pxempjt(\x_{jt})
     \frac{\pthetazxj(\z_t| \x_{jt})}{\pthetazj(\z_t)}
     \right) \pthetaz(\setZ) \,,
\end{equation}
Each recognition factor is parametrised by a neural network $\theta_j$ that outputs the natural parameters $\etaxjn$ of a multivariate normal distribution given input $\x_{jt} \nn$
and we recall that
\begin{equation}
  \pthetazj(\z_t) = \frac1N \sum_{n=1}^N 
  \pthetazxj \left( \z_t| \x_{jt}\nn \right) \,. 
\end{equation}
The prior on $\setZ$ comprises independent Gaussian Process priors over each latent dimension $\setZ_k = \{z_{kt} : t = 1\dots T\}$
\begin{equation}
    \pthetaz(\setZ) = \prod_{k=1}^{K} p_k(\setZ_k) \text{ ; }
    p_k(\cdot) = \mathcal{GP}(0, \kappa_k(\cdot,\cdot))\,,
\end{equation}
\subsection{Variational Distribution and inducing points}
We use sparse variational GP approximations \citep{titsias2009variational} to improve scalability of RP-GPFA. The model is augmented with $M$ inducing points (IP) for each latent dimension ($k=1\dots K$) and each observation ($n=1\dots N$). This smaller set ($M < T$) of fictitious measurements is optimised to efficiently represent function evaluations. For simplicity, IP are defined at fixed and shared locations $\tau = \left[\tau_1, \dots, \tau_M \right]^\top$. We denote them
\begin{equation}
    \setU\nn = \left[ 
    \setU_1\nn, \dots, \setU_K\nn
    \right] \sim M \times K \,.
\end{equation}
Given an observation $n$, the variational distribution writes
\begin{equation}
    q\left(\setU\nn, \setZ\nn \right) = \prod_{k=1}^K 
    q\left(\setU_k\nn, \setZ_k\nn \right) \,.
\end{equation}
In practice, we only need the marginals over inducing points and latents. The former are optimised numerically and denoted
\begin{equation}
q \left(
    \setU_k\nn
    \right) =
    \mathcal{N}
    \left(
    \mu_k\nn, \Sigma_k\nn
    \right) \,.
\label{eq:variational}
\end{equation}
For the latter, we simplify inference by adopting the form
\begin{equation}
    q\left(\setU_k\nn, \setZ_k\nn \right) = 
    p_k \left(
    \setZ_k\nn | \setU_k\nn
    \right)
    q \left(
    \setU_k\nn
    \right) \,,
\end{equation}
which gives closed form expression for
\begin{equation}
    q\left(z_{k,t}\nn \right) = \mathcal{N}
    \left(
    m_{k,t} \nn, S_{k,t} \nn 
    \right) \,.
\end{equation}
Indeed, we denote $\kappa_k^\tau = \kappa_k(\tau, \tau)$ and use the law of total expectations to obtain
\begin{equation}
    m_{kt}\nn = \mathbb{E}_u (\mathbb{E}_{z|u}(z)) = \kappa_k(t, \tau) \kappa_k^{\tau -1} \mu_k\nn
\end{equation}
and
\begin{equation}
\begin{split}
    s_{kt}\nn =& \mathbb{V}_u (\mathbb{E}_{z|u}(z)) + \mathbb{E}_u (\mathbb{V}_{z|u}(z)) 
    = \kappa_k(t, \tau) 
    \left(
    \kappa_k^{\tau -1} \Sigma_k\nn \kappa_k^{\tau -1}
    -\kappa_k^{\tau -1} 
    \right) \kappa_k(\tau, t)
    + \kappa_k(t,t) \,.
\end{split}
\end{equation}
We gather those centred moments in the $K$ dimensional vector $\m\nn_t$ and diagonal matrix $S_t\nn$.

\subsection{Variational Free Energy}

The free energy is given by
\begin{equation}
\begin{split}
    \log \pthetaXN{\setX} &= \mltintdx[2]{d\setZ\,d\setU} \pthetaXN{\setX, \setZ, \setU} \\
    & \geq \mltintdx[2]{d\setZ\, d\setU} q\left(\setZ, \setU \right)
    \log \frac{\pthetaXN{\setX | \setZ} \pthetaz(\setZ | \setU) \pthetaz(\setU)}{\pthetaz(\setZ | \setU) q(\setU)} \\
    &=
    \langle \log \pthetaXN{\setX | \setZ} \rangle_{q(\setZ, \setU)} - 
    \KL[\Big]{q (\setU)}{\pthetaz(\setU)}
    = \mathcal{F} 
    \end{split}
\end{equation}
The KL divergence between the variational and the prior distribution over IP is closed form and can be broken down to
\begin{equation}
    \KL[\Big]{q (\setU)}{P(\setU)}
    = \sum_{n,k} \KL[\Big]{q (\setU_k\nn)}{p_k(\setU_k\nn)}
\end{equation}
The remaining term has the RPM form
\begin{equation}
\begin{split}
    \big\langle \pthetaXN{\setX | \setZ} \big\rangle_{q(\setZ, \setU)} 
    = NJT \log \frac{1}{N} + 
     \sum_{njt} \big\langle \log \pthetazxj(\z_t| \x_{jt}\nn) \big\rangle_{q \left( \z_t\nn \right)} 
     -  \big\langle \log \pthetazj(\z_t) \big\rangle_{q \left( \z_t\nn \right)} \,,
\end{split}
\end{equation}
and is estimated by using one of the inference methods described above. Finally, the free energy (or its lower bound) is optimised with respect to the kernel parameters, the inducing point variational distributions, and the recognition networks (and the auxiliary factors) using Adam. When necessary, we ensure the validity of $\hat f_j\nn$ by soft-thresholding the eigenvalues of the natural parameters.

\subsection{RP-GPFA Experiments}

In all RP-GPFA experiments, the recognition networks consisted in at least two fully connected hidden layers of size 50. When input included image frames, they were preceded by two convolutional layer with max pooling. All layers were followed by Rectified Linear Unit (ReLU) and trained with Adam.
\subsubsection*{Bouncing Balls}
In Bouncing ball experiments, the latent was generated with a randomly initialised two dimensional oscillating linear system from which we extracted the first components. We fixed the number of observation to $N=50$, the number of time points to $T=50$, and used $M=20$ inducing points. As described in the main text, in the textured experiment, the stochastic mapping from latent to observation is defined such that the mean and variance of pixel intensity is independent of the latent position (respectively fixed to $0$ and $1$). This is achieved with a mixture of Gaussian distributions with fixed variance, but whose weights and position depend on the latent. 
\subsubsection*{Multi-factorial experiment}
In the multi-factorial experiment, three independent agents are placed in a bounded environment where they work to move inanimate blocks to a designated target. Once the task is complete, the arena ground colour changes. Observations of the agents' locations are collected in the form of 3D-rendered image frames of the entire environment and noisy range-finding sensor data giving the distances of one of the agents from the four corners of the room. Sensor noise is modelled as additive Gaussian with zero mean and variance $0.1$.
Renderings and trajectories were generated using Unity Machine Learning Agents Toolkit \citep{juliani2018unity}. We used $N=50$ observations of length $T=200$ and $M=40$ inducing points.

\section{Compute time}
\label{sec:supp-compute}

Compute time for the RPM experiments was competitive with the baseline comparison methods for full-batch training with both discrete and continuous latents.  

\begin{figure}

  \begin{tikzpicture}[every node/.style={font={\small\sf}}]
    \node[inner sep=0pt] (peerfig) 
        {\includegraphics[width=0.44\linewidth,height=4cm, trim=70 110 120 100,clip]{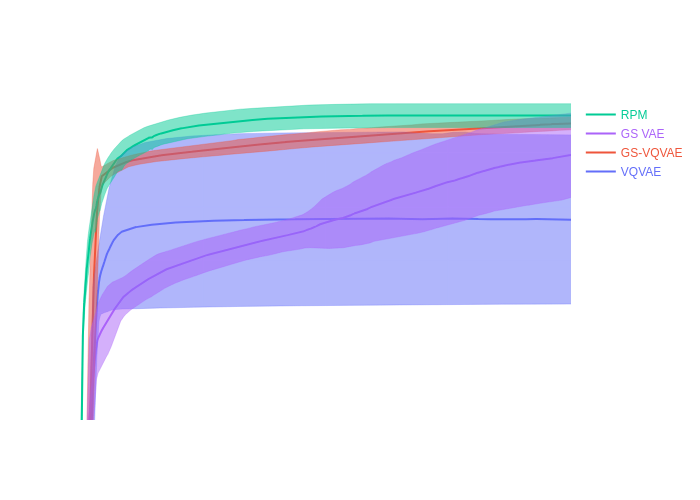}};

    \begin{scope}[fit to node=peerfig]
      \node at (0.5,1.03) {Peer supervision};
      \node[rotate=90,above] at (0,0.4) 
        {$\displaystyle \frac{\eff_{\textrm{max}} - \eff}{\eff_{\textrm{max}} - \eff_{\textrm{min}}}$};
      \node[left=-0.2em] at (0,0.95) {0};
      \draw[very thin] (0,1) -| (0.02,0);
      \node[below] at (0.5,0) {time (min)};
      \draw[very thin] (0.02,-0.2ex) |- (0.5,0) -| (1,-0.2ex);
      \node[below=-0.5ex] at (0.03,0) {0};
      \node[below=-0.5ex] at (0.98,0) {800};
      \node[matrix of nodes] at (0.7,0.12) {
          |[right,text={rgb:red,0;green,204;blue,150}]| RPM &
          |[right,text={rgb:red,99;green,110;blue,250}]| VQVAE\\
          |[right,text={rgb:red,239;green,85;blue,59}]| GS-VQVAE &
          |[right,text={rgb:red,200;green,99;blue,200}]| GS VAE \\
          };
    \end{scope}

    \node[inner sep=0pt, right=0mm of peerfig] (ballfig) 
        {\includegraphics[width=0.44\linewidth,height=4cm, trim=70 42 155 15,clip]{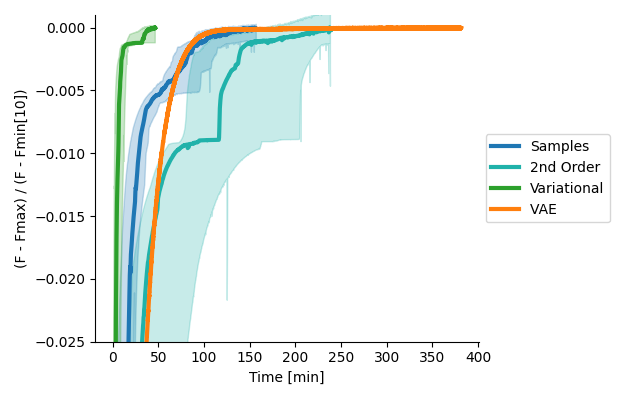}};

    \begin{scope}[fit to node=ballfig]
      \node at (0.5,1.03) {Bouncing ball};
      \node[below] at (0.5,0) {time (min)};
      \draw[very thin] (0.05,-0.2ex) |- (0.5,0) -| (0.94,-0.2ex);
      \node[below=-0.5ex] at (0.07,0) {0};
      \node[below=-0.5ex] at (0.94,0) {300};

      \matrix [matrix of nodes, right, inner sep=2pt, right delimiter=\}] 
        (rpm legend)
        at (0.7, 0.6) {
          |[right,text=green!50!black]| variational\\
          |[right,text=blue!40!black!80!white]| reparam\vphantom{l}\\
          |[right,text=cyan!80!black]| 2nd order\\
        };
     \node [right = 1em of rpm legend] {RPM};
     \node[below right = 1ex and 0pt of rpm legend.south west,text=orange!80!black]
        {sGPVAE};

    \end{scope}
  \end{tikzpicture}

  \caption{Relative free energy vs clock time for MNIST peer supervision (left) and a Bouncing ball data set (right).  Free energy is shown relative to the highest value at convergence (rather than on an absolute scale) to emphasise relative timing.  In both cases the RPM approaches converged to higher absolute values of free energy than alternatives.}
  \label{fig:timing}

\end{figure}

\cref{fig:timing} shows wall-clock comparisons for MNIST peer supervision, and for a bouncing ball (with noise better matched for sGPVAE so that it converges to a non-trivial value).  Learning curves are scaled vertically to emphasise relative timing.  The RPM always converged to a higher value of free energy on an absolute scale.  

We used 20 samples per latent in reparametrisation to ensure that the compute time was comparable.  Note that although the variational method is fast here, the current implementation scales poorly with GP dimension.  

\section{Code}
\label{sec:supp-code}

Implementation and code of all discrete and continuous latent experiments can be found at \url{https://github.com/gatsby-sahani/rpm-aistats-2023}


%% file: figures/peer_supervision/peer_supervision_confusion.tikz
\usetikzlibrary{plotmarks}

\colorlet{axiscol}{black!50}
\colorlet{gridcol}{black!10}

\colorlet{rpmcol}{blue!80!black!70}
\colorlet{vqcol}{orange!80!black!70}
\colorlet{gsvqcol}{green!80!black!70}
\colorlet{gscol}{purple!80!black!70}

\begin{tikzpicture}[
    ticklabel/.style={font={\small\sf},text=axiscol},
    legend/.style={font={\small\sf},text=axiscol},
    confusion/.style={text width=2.3cm,align=center,inner sep=0pt},
    every mark/.add style={thick,scale=1}{}
    ]

  \begin{scope}[x={(15cm,0)}, y={(0,7.5cm)}]

    \draw[thin,gridcol] (1.7,0) grid (2.4,1);

    \draw[thick,axiscol] (1.7,0) -- (1.75,0)
      +(-135:1mm) -- +(45:1mm)
      ++(0.01,0)
      +(-135:1mm) -- +(45:1mm)
      ++(0,0) -- (2.4,0);

    \draw[thick,axiscol] (1.7,0) -- (1.7,1);

    \node at (2.1,-0.05) [below,ticklabel] {entropy of the average posterior};
    
    \node at (1.625,0.5) [below,ticklabel,rotate=90] {accuracy};

    \node at (1.7,0) [below,ticklabel] {0};
    \foreach \x in {1.8,2.0,...,2.4} \node at (\x,0) [below,ticklabel] {\x};
    \foreach \y in {0,0.2,0.4,0.6,0.8,1.0} \node at (1.7,\y) [left,ticklabel] {\y};
    

    \begin{scope}
      \clip (1.8,0) rectangle (2.4,1);
      
      \draw[rpmcol]
      plot[only marks,mark=o] file{figures/peer_supervision/entropy_vs_accuracy.rpm.train}
      plot[only marks,mark=*] file{figures/peer_supervision/entropy_vs_accuracy.rpm.test};

      \draw[vqcol]
      plot[only marks,mark=o] file{figures/peer_supervision/entropy_vs_accuracy.vq.train}
      plot[only marks,mark=*] file{figures/peer_supervision/entropy_vs_accuracy.vq.test};

      \draw[gsvqcol]
      plot[only marks,mark=o] file{figures/peer_supervision/entropy_vs_accuracy.gsvq.train}
      plot[only marks,mark=*] file{figures/peer_supervision/entropy_vs_accuracy.gsvq.test};

      \draw[gscol]
      plot[only marks,mark=o] file{figures/peer_supervision/entropy_vs_accuracy.gs.train}
      plot[only marks,mark=*] file{figures/peer_supervision/entropy_vs_accuracy.gs.test};
    \end{scope}

    \begin{scope}[shift={(1.7,0)}]
      \clip (-0.1,0) rectangle (0.2,1);
      \draw[orange!80!black!70]
      plot[only marks,mark=o] file{figures/peer_supervision/entropy_vs_accuracy.vq.train}
      plot[only marks,mark=*] file{figures/peer_supervision/entropy_vs_accuracy.vq.test};
    \end{scope}

    \path plot[only marks,mark=o] coordinates{(1.72,0.95)} node[right,legend] {train}
          plot[only marks,mark=*] coordinates{(1.8,0.95)} node[right,legend] {test}
          plot[only marks,mark=*,every mark/.append style={rpmcol}] coordinates{(1.72,0.9)} node[right,legend] {RPM}
          plot[only marks,mark=*,every mark/.append style={vqcol}] coordinates{(1.72,0.85)} node[right,legend] {VQVAE}
          plot[only marks,mark=*,every mark/.append style={gsvqcol}] coordinates{(1.72,0.8)} node[right,legend] {GS-VQVAE}
          plot[only marks,mark=*,every mark/.append style={gscol}] coordinates{(1.72,0.75)} node[right,legend] {GS-VAE}
          ;

    \node (rpm1 conf) at (2.1,1) [above = 1em,confusion]
          {\includegraphics[width=\linewidth,trim=100 50 200 50,clip]{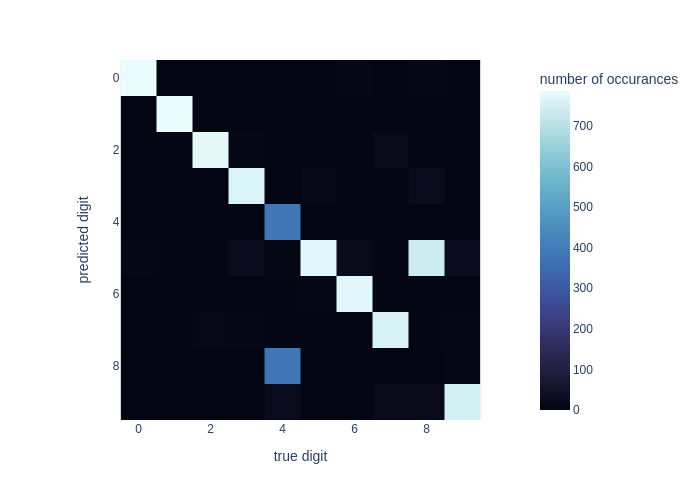}}
          edge[->] (2.171,0.683);

    \node (rpm2 conf) at (2.3,1) [above = 1em,confusion]
          {\includegraphics[width=\linewidth,trim=100 50 200 50,clip]{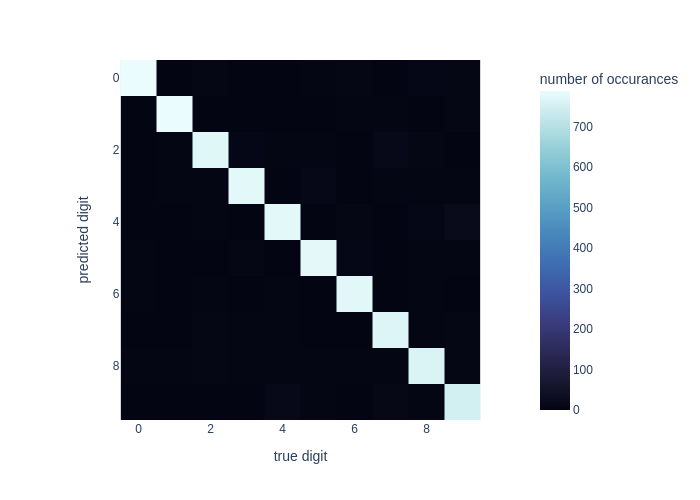}}
          edge[->] (2.303,0.965);

    \node (gsvqvae conf) at (2.4,1) [below right = 0pt and 1em,confusion]
          {\includegraphics[width=\linewidth,trim=100 50 200 50,clip]{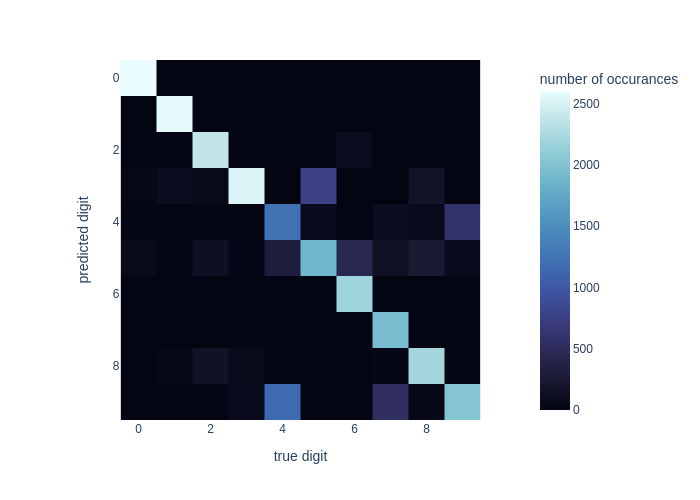}}
          edge[->] (2.278,0.812);

    \node (vqvae conf) [below = 1em of gsvqvae conf,confusion]
          {\includegraphics[width=\linewidth,trim=100 50 200 50,clip]{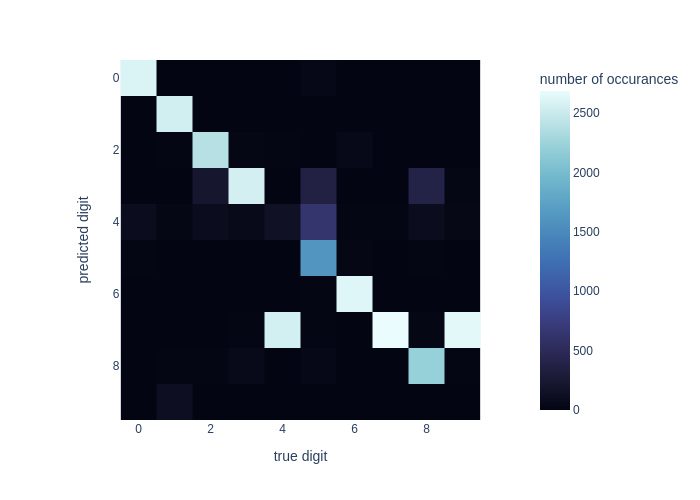}}
          edge[->] (2.065,0.708);

    \node (gsvae conf) [below = 1em of vqvae conf,confusion]
          {\includegraphics[width=\linewidth,trim=100 50 200 50,clip]{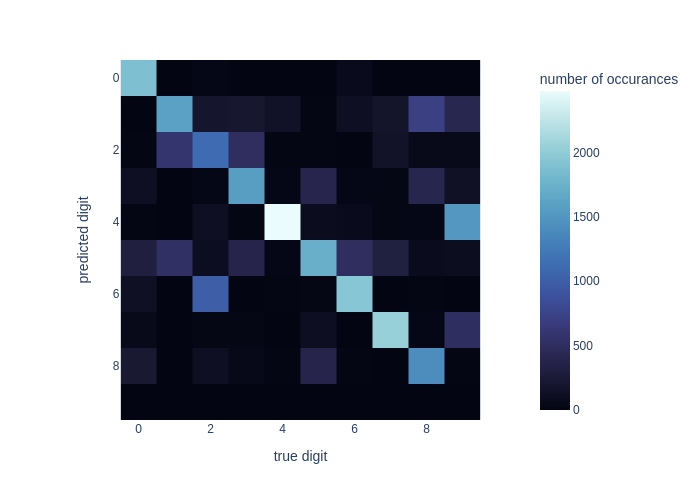}}
          edge[->] (2.168,0.575);
          
    \path[fit to node=rpm1 conf]
      (0.1,0) node[ticklabel] {0}
      (0.9,0) node[ticklabel] {9}
      (1,0.1) node[ticklabel] {9}
      (1,0.9) node[ticklabel] {0};

    \path[fit to node=rpm2 conf]
      (0.1,0) node[ticklabel] {0}
      (0.9,0) node[ticklabel] {9}
      (1,0.1) node[ticklabel] {9}
      (1,0.9) node[ticklabel] {0};

    \path[fit to node=gsvqvae conf]
      (0.1,0) node[ticklabel] {0}
      (0.9,0) node[ticklabel] {9}
      (1,0.1) node[ticklabel] {9}
      (1,0.9) node[ticklabel] {0};

    \path[fit to node=vqvae conf]
      (0.1,0) node[ticklabel] {0}
      (0.9,0) node[ticklabel] {9}
      (1,0.1) node[ticklabel] {9}
      (1,0.9) node[ticklabel] {0};

    \path[fit to node=gsvae conf]
      (0.1,0) node[ticklabel] {0}
      (0.9,0) node[ticklabel] {9}
      (0.5,0) node[below,ticklabel] {true digit}
      (1,0.1) node[ticklabel] {9}
      (1,0.9) node[ticklabel] {0}
      (1,0.5) node[below,rotate=90,ticklabel] {predicted digit};

  \end{scope}


\end{tikzpicture}